\documentclass[letterpaper]{article} % DO NOT CHANGE THIS
\usepackage{aaai23}  % DO NOT CHANGE THIS
\usepackage{times}  % DO NOT CHANGE THIS
\usepackage{helvet}  % DO NOT CHANGE THIS
\usepackage{courier}  % DO NOT CHANGE THIS
\usepackage[hyphens]{url}  % DO NOT CHANGE THIS
\usepackage{graphicx} % DO NOT CHANGE THIS
\usepackage{natbib}  % DO NOT CHANGE THIS AND DO NOT ADD ANY OPTIONS TO IT
\usepackage{caption} % DO NOT CHANGE THIS AND DO NOT ADD ANY OPTIONS TO IT
\usepackage{amsmath,amssymb,amsfonts,amssymb,multirow}
\usepackage{graphicx}
\usepackage{xcolor}
\usepackage{subfigure}
\usepackage{url,setspace}
\usepackage{colortbl}
\usepackage[normalem]{ulem}
\useunder{\uline}{\ul}{}
\usepackage[boxed,ruled,vlined,linesnumbered]{algorithm2e}
\usepackage{wrapfig}
\usepackage{algpseudocode}
\usepackage{multirow}
\usepackage{multicol}
\usepackage{makecell}
\usepackage{textcomp}
\usepackage{enumitem}
\usepackage{booktabs}
\usepackage{bm}
\usepackage{amsmath,amsfonts,bm}

\def\eqref#1{equation~\ref{#1}}
\def\1{\bm{1}}

\def\vc{{\bm{c}}}

\def\vh{{\bm{h}}}

\def\vm{{\bm{m}}}

\def\vx{{\bm{x}}}

\def\vz{{\bm{z}}}

\def\mA{{\bm{A}}}

\def\mD{{\bm{D}}}

\def\mH{{\bm{H}}}
\def\mI{{\bm{I}}}

\def\mW{{\bm{W}}}
\def\mX{{\bm{X}}}
\def\mY{{\bm{Y}}}
\def\mZ{{\bm{Z}}}

\DeclareMathAlphabet{\mathsfit}{\encodingdefault}{\sfdefault}{m}{sl}
\SetMathAlphabet{\mathsfit}{bold}{\encodingdefault}{\sfdefault}{bx}{n}

\def\gE{{\mathcal{E}}}

\def\gG{{\mathcal{G}}}

\def\gV{{\mathcal{V}}}

\def\sR{{\mathbb{R}}}

\let\oldnl\nl% Store \nl in \oldnl
\newcommand{\nonl}{\renewcommand{\nl}{\let\nl\oldnl}}% Remove line number for one line

\usepackage{amsthm}

\newtheorem{fact}{Fact}
\newtheorem{prop}{Proposition}

\newcommand{\ie}{\textit{i}.\textit{e}., }
\newcommand{\eg}{\textit{e}.\textit{g}., }

\SetKwInOut{Parameter}{Parameter}

\title{ImGCL: Revisiting Graph Contrastive Learning on\\ Imbalanced Node Classification}

\author{
Liang Zeng\textsuperscript{\rm 1},
Lanqing Li\textsuperscript{\rm 2}\thanks{Corresponding authors.},
Ziqi Gao\textsuperscript{\rm 3},
Peilin Zhao\textsuperscript{\rm 2},
Jian Li\textsuperscript{\rm 1}\footnotemark[1]
}
\affiliations{
\textsuperscript{\rm 1}Institute for Interdisciplinary Information Sciences (IIIS), Tsinghua University\\
\textsuperscript{\rm 2}Tencent AI Lab\\
\textsuperscript{\rm 3}Hong Kong University of Science and Technology \\
zengl18@mails.tsinghua.edu.cn, \{lanqingli, masonzhao\}@tencent.com,\\ zgaoat@connect.ust.hk, lijian83@mail.tsinghua.edu.cn
}
\begin{document}
\maketitle

\begin{abstract}
Graph contrastive learning~(GCL) has attracted a surge of attention due to its superior performance for learning node/graph representations without labels. However, in practice, the underlying class distribution of unlabeled nodes for the given graph is usually imbalanced. This highly imbalanced class distribution inevitably deteriorates the quality of learned node representations in GCL. Indeed, we empirically find that most state-of-the-art GCL methods cannot obtain discriminative representations and exhibit poor performance on imbalanced node classification. Motivated by this observation, we propose a principled GCL framework on \underline{Im}balanced node classification~(ImGCL), which automatically and adaptively balances the representations learned from GCL without labels. Specifically, we first introduce the online clustering based \emph{progressively balanced sampling}~(PBS) method with theoretical rationale, which balances the training sets based on pseudo-labels obtained from learned representations in GCL. We then develop the node centrality based PBS method to better preserve the intrinsic structure of graphs, by upweighting the important nodes of the given graph. Extensive experiments on multiple imbalanced graph datasets and imbalanced settings demonstrate the effectiveness of our proposed framework, which significantly improves the performance of the recent state-of-the-art GCL methods. Further experimental ablations and analyses show that the ImGCL framework consistently improves the representation quality of nodes in under-represented~(tail) classes.
\end{abstract}

\section{Introduction}
\label{sec:introduction}
Recently, graph contrastive learning~(GCL) has become the \emph{de facto standard} for self-supervised learning on graphs~\cite{zhu2021empirical} due to its superior performance as compared to the supervised counterparts.~\cite{thakoor2021bootstrapped, bielak2021graph, you2020graph, zhu2020graph}. Inheriting the advantage of self-supervised learning, GCL frees the model from the reliance on label information in the graph domain, where labels can be costly and error-prone in practice~\cite{yang2020rethinking} while unlabeled/partially labeled data is prevalent, such as fraudulent user detection~\cite{kumar2018rev2} and molecular property prediction~\cite{ma2020multi}. 
Typically, most GCL methods first construct multiple graph views via stochastic augmentation functions on the input graph and then learn discriminative representations by maximizing the representation consistency between two views.

\begin{figure}[t]
    \includegraphics[width=1.\linewidth]{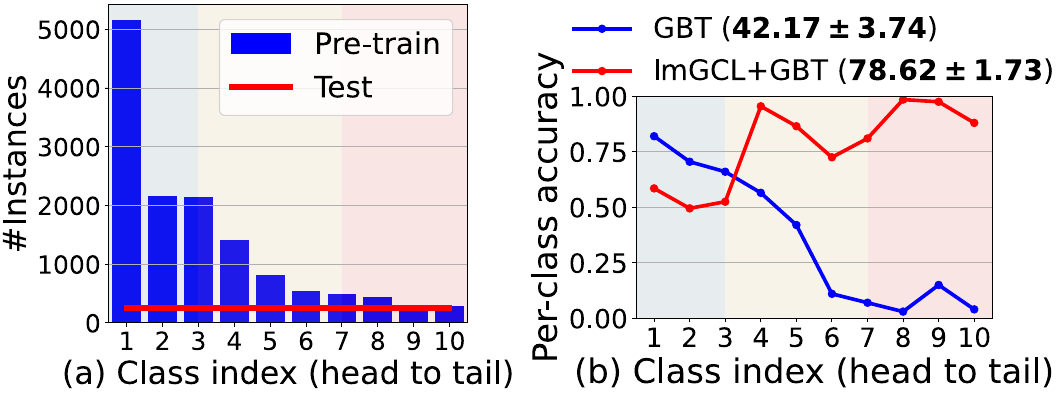}
    \vspace{-0.3cm}
    \caption{(a) Class distribution of the Amazon-Computers dataset sorted in decreasing order, where pre-training sets are highly \emph{imbalanced} but testing sets are \emph{balanced}. (b) Compared to the GBT~\cite{bielak2021graph} baseline model on Amazon-computers, ImGCL+GBT substantially outperforms GBT on the overall accuracy. On three splits~(\emph{head/middle/tail}) depicted in different colors~(\emph{blue/yellow/red}), ImGCL+GBT improves the performance on middle and tail classes by a large margin with a slight sacrifice of accuracy on head classes.}
    \label{fig:illustrate}
    \vspace{-0.5cm}
\end{figure}
Despite the prevalence and effectiveness of such methods, existing GCL methods mostly assume the datasets are carefully curated and well balanced across classes. 
However, unlabeled graph data randomly gathered in the wild often exhibits a highly imbalanced class distribution~\cite{barabasi1999emergence,liu2021tail} and thus implicitly deteriorates the quality of learned representations in GCL methods~\cite{jiang2021self}. For instance, Fig.~\ref{fig:illustrate}(a) illustrates the class distribution sorted in decreasing order of the Amazon-Computers dataset~\cite{shchur2018pitfalls}, a network of co-purchase goods containing 10 classes. 
In the pre-training set, we can clearly find that a small fraction of classes take up massive samples~(\emph{a.k.a.}, head classes) and the rest of classes are assigned only a few samples~(\emph{a.k.a.}, tail classes). 
Note that the highly \emph{imbalanced} property of label information within the training data is unknown to the GCL methods. 
This important property implicitly exists and is largely ignored by the current GCL methods~\cite{zhu2021empirical}. However, to impose a fair evaluation metric, on imbalanced node classification, the testing set is \emph{balanced} across all classes. Shown as the blue line of Fig.~\ref{fig:illustrate}(b), we adopt one of the state-of-the-art GCL methods---GBT~\cite{bielak2021graph}---to conduct experiments on imbalanced node classification. 
We can find that the baseline GBT model obtains very poor results, especially on the under-represented middle and tail classes, which naturally spurs a question:

\emph{How to improve the representation learning of GCL on highly imbalanced node classification?}

Recent works related to this question~\cite{he2021rethinking, kang2020exploring} explore balanced feature spaces to learn powerful representations not just for head classes but also for tailed classes.
\citet{kang2019decoupling} introduce PBS method to innovate imbalanced representation learning, achieving remarkable success by decoupling the learning procedure into a representation learning stage and a classification stage. 
However, this method is impractical for the GCL setting since it requires knowing labels. This dilemma motivates us to explore how to implicitly obtain label information to improve node representations in the traditional GCL setting.

In this paper, we present a principled GCL framework on \underline{Im}balanced node classification (ImGCL), to automatically and adaptively balance the representations learned from GCL without ground truth labels.
To perform class-balanced re-sampling, ImGCL obtains pseudo-labels via online clustering of learned representations in GCL.
Moreover, we propose the node centrality based PBS method tailored for the graph domain, which assigns a higher probability to retain nodes with high node centrality scores when down-sampling the head class nodes in online clustering based PBS. 
This scheme is able to guide the model to learn node representations with higher node centrality scores, which is considered more important by having abundant structural connectivity when performing message passing.
We also provide theoretical insight into the PBS method. 
Furthermore, existing GCL models can be seamlessly incorporated into the proposed ImGCL framework in a plug-and-play manner.
In short, our main contributions can be summarized as follows:
\begin{itemize}[leftmargin=10pt]
    \item \emph{New problem and insights:} we introduce a practically important but under-explored problem, namely graph contrastive learning on imbalanced node classification. We empirically identify that the recently proposed GCL methods are \emph{vulnerable} to node class imbalance and result in large performance degradation.
    \item \emph{New principled framework:} we propose a novel ImGCL framework, which utilizes the node centrality based PBS method. ImGCL automatically and adaptively balances the representations learned from GCL without knowing labels. Moreover, existing GCL models can be seamlessly incorporated into our framework.
    \item \emph{Convincing empirical results:} we conduct comprehensive experiments to show that the ImGCL framework achieves superior performance compared with the recently proposed GCL methods on imbalanced node classification. Extensive ablation studies also demonstrate that the ImGCL framework improves the representations of the under-represented~(middle and tail) classes.
\end{itemize}
\section{Background}
\label{sec:background}
\begin{figure*}[t]
    \centering
    \includegraphics[width=0.92\linewidth]{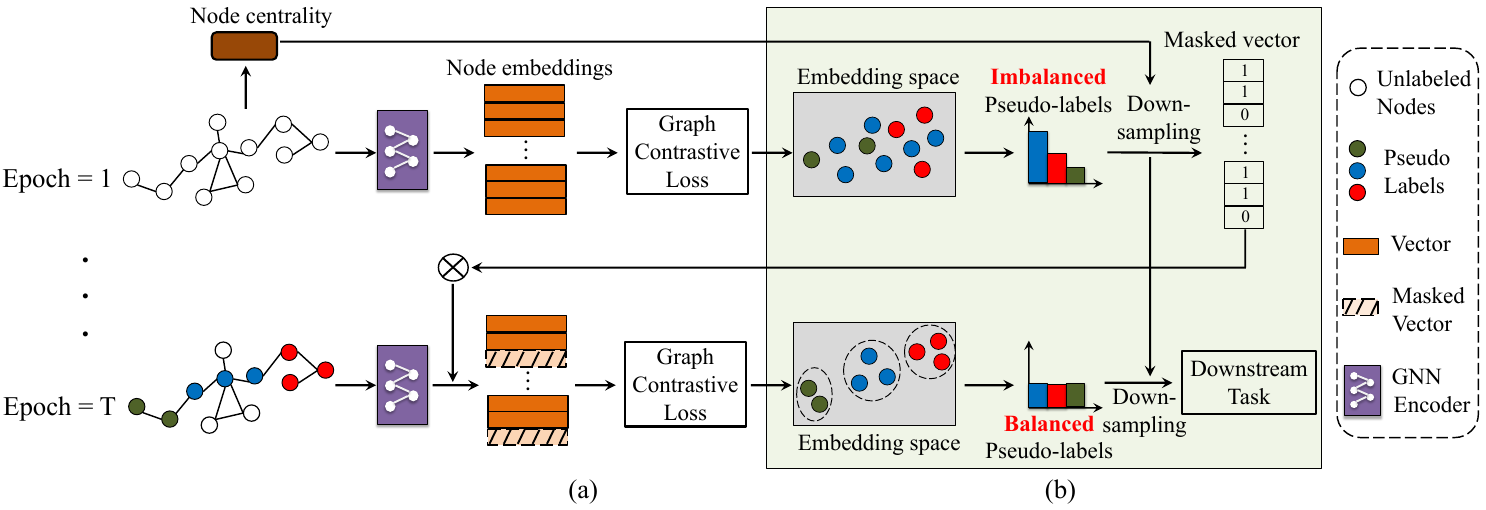}
    % \vspace{-0.5em}
    \caption{Overview of the proposed ImGCL framework. (a) Graph contrastive learning~(GCL) methods take the graph as input and produce the embeddings of each node. (b) Node centrality based progressively balanced sampling~(PBS) method automatically and adaptively balances the representations learned from GCL without knowing the labels.}
    \label{fig:framework}
    \vspace{-1em}
\end{figure*}

Let $\gG=\left(\gV, \gE, \mX\right)$ denote a graph, where $\gV = \left\{v_i\right\}_{i=1}^N$ is a set of $N$ nodes, and $\gE \subseteq \gV \times \gV$ is a set of edges between nodes. $\mX = \left[\vx_1, \vx_2, \cdots, \vx_N\right]^{T}$ $\in \sR^{N \times d}$ represents the node feature matrix and $\vx_i \in \sR^d$ is the feature vector of node $v_i$, where $d$ is the feature dimension. The adjacency matrix $\mA \in \{0,1\}^{N \times N}$ is defined by $\mA_{i,j}=1$ if $\left(v_i, v_j\right) \in \gE$ and $0$ otherwise. More detailed discussions about related works can be found in the appendix.
% The degree matrix $\mD \in \mathbb{R}^{N\times N}$ is a diagonal matrix defined by $\mD_{ii} = \sum_{j} \mA_{ij}$. 
% Here we briefly introduce relevant lines of research and more detailed discussions about related works can be found in the appendix.
% In this work, we focus on undirected and unweighted graphs. However, it is straightforward to extend this work to the directed and weighted graph by adopting the corresponding GCL methods.

\paragraph{Graph Contrastive Learning~(GCL).} Given an input graph, GCL aims to learn effective graph/node representations that can be transferred to downstream tasks by constructing positive and negative sample pairs~\cite{thakoor2021bootstrapped, bielak2021graph, zhang2021canonical, xu2021infogcl, velivckovic2018deep, sun2019infograph, hassani2020contrastive}.
GCL maximizes the representation consistency between two augmented views of the input graph via contrastive loss in the latent space~\cite{zhu2021empirical, you2020graph}. 
We first perform stochastic \emph{graph augmentations} to generate multiple views of the original graph. Specifically, we utilize two augmentation functions $t_1, t_2 \sim \mathcal{T}$ to generate graph views $\widetilde{\gG}_1=(\widetilde{\mX}_1,\widetilde{\mA}_1)=t_1(\gG)$ and $\widetilde{\gG}_2=(\widetilde{\mX}_2,\widetilde{\mA}_2)=t_2(\gG)$, where $\mathcal{T}$ is the set of graph augmentation functions, such as node dropping, edge perturbation, and subgraph sampling~\cite{zhu2020graph}. We then obtain node representations for the two graph views via the~(parameter shared) GNN encoder $f(\cdot)$, denoted by $\mZ = f(\widetilde{\mX}_1,\widetilde{\mA}_1)$ and $\mZ' = f(\widetilde{\mX}_2,\widetilde{\mA}_2)$ respectively. 
Given the latent representations, we optimize the parameters of the GNN encoder by a pre-defined contrastive loss. 
For any node $u$ in $\widetilde{\gG}_1$, we aim to score the positive pairs ($u$, $u^+$) higher compared to other negative pairs ($u$, $u^-$). Typically, the negative samples $u^-$ are sampled from other nodes of the augmented graph views $\widetilde{\gG}_2$ in the same batch. The commonly-used InfoNCE loss~\cite{chen2020simpleframework} can be defined as: 
\begin{equation}
    \mathcal{L}_{NCE}(u) = -\log \frac{s\left(\vz_u,\vz_{u^+},\tau \right)}{s\left(\vz_u,\vz_{u^+},\tau \right)+\sum_{u^{-} \neq u}s\left(\vz_u,\vz_{u^-},\tau\right)} ,
\label{equ:1}
\end{equation}
where $s\left(\vz_u,\vz_{u^+},\tau \right)$ indicates the similarity between node representations of positive pairs, while $s\left(\vz_u,\vz_{u^-},\tau\right)$ is the similarity between negative pairs.
% where $\vz_u$ and $\vz_{u^+}$ denote the node representation vector of positive pair $(u, u^+)$ in $\widetilde{\gG}_1$, respectively. $\vz_{u^-}$ denotes the node representation of node $u^-$ in $\widetilde{\gG}_2$. 
$s$ is a contrasting function to measure the similarity between two node representations, which is typically defined as: $s(\vz_u, \vz_v, \tau) = \exp(\vz_u \cdot \vz_v / \tau)$. $\tau$ represents the temperature hyper-parameter.
% and $N$ denotes the batch size. 
% The similarity function is typically defined as: $s(\vz_u, \vz_v, \tau) = \exp(\vz_u \cdot \vz_v / \tau)$.

\paragraph{Imbalanced Learning.} Imbalanced learning seeks to learn a model from \emph{the training set with an imbalanced class distribution}, where head classes take up the vast majority of samples and tail classes occupy only a few samples, and generalize well on \emph{a balanced testing set}~\cite{kang2019decoupling, zhang2021deep, zhang2021bag, liu2019large, li2022imdrug}. 
For a $K$-way node classification problem, let $\{v_i, y_i\}_{i=1}^N$ be an imbalanced training set.
The total number of the training set over $K$ classes is $N=\sum_{k=1}^K N_k$, where $N_k$ denotes the number of samples in class $k$. Let $\bm{\pi}$ be the vector of label frequencies, where $\pi_k=N_k/N$ denotes the label frequency of class $k$. Without loss of generality, we assume that the classes are sorted by $\pi_k$ in a descending order (\ie if the class index $i < j$, then $N_{i} \geq N_{j}$, and $N_1 \gg N_K$). 
We denote by $N_1/N_K$ the imbalance ratio of the dataset. 
% Note that in imbalanced learning, the training dataset is imbalanced~(long-tailed) to mimic the power-law class distribution of real-world data collection~\cite{barabasi1999emergence}, but the testing dataset is balanced assuming equality of all classes.
Conventionally, evaluations on imbalanced learning report statistics for the head, middle, tail, and overall classes separately.
% the overall performance on \emph{all classes} and the performance for \emph{the head, middle, and tail classes} are usually reported.
\section{ImGCL: The Proposed Framework}
\label{sec:imgcl}
In this section, we first introduce the progressively balanced sampling~(PBS) method, which is impractical in our self-supervised setting since it requires knowing the labels. Motivated by the property of GCL, we generate pseudo-labels by clustering the node representations.
Finally, we utilize the proposed node centrality based PBS method to adaptively attend to 'important' nodes of the given graph during the down-sampling phase. The overall framework of ImGCL is shown in Fig.~\ref{fig:framework}.

\subsection{Progressively Balanced Sampling~(PBS)}
\label{sec:pbs}
\paragraph{Sampling Strategies.} As suggested in~\cite{he2021rethinking}, the probability $p_k$ of sampling a node for the given graph from the class $k$ is defined as:
\begin{equation}
    p_k = \frac{N_k^q}{\sum_{i=1}^K N_i^q} \;,
    \label{equ:sampling}
\end{equation}
where $q \in [0,1]$. Different sampling strategies have different specific values of $q$.

\paragraph{PBS~\cite{kang2019decoupling}.} In imbalanced learning, the testing dataset is \emph{balanced} whereas the training dataset is highly \emph{imbalanced}. 
A single data sampling strategy which fits only one case, the balanced dataset~($p_{k}^{M}=\frac{1}{K}$ by setting $q=0$ in Eq.~\ref{equ:sampling}) or the imbalanced dataset~($p_{k}^{R} = \frac{N_k}{\sum_{i=1}^K N_i}$ by setting $q=1$ in Eq.~\ref{equ:sampling}), cannot account for the class distribution shift between the training and testing datasets. 
Thus, in order to learn high-quality representations from the imbalanced dataset, we adopt two data samplers with adaptive sampling strategies known as decoupled training in long-tailed learning literature~\cite{zhang2021deep}. 
Formally, at training step $t$, data are sampled according to a linear combination of the random and mean strategies, controlled by a parameter $\alpha \in [0,1]$. 
Therefore, the probability of sampling a node from the class $k$ is given by:
\begin{equation}
\begin{aligned}
    p_k^{\text{PB}} &= \alpha*p_{k}^{R} + (1-\alpha)*p_{k}^{M} \\
    &=\alpha*\frac{N_k}{\sum_{i=1}^K N_i} + (1-\alpha)*\frac{1}{K}.
\end{aligned}
\label{eq:3}
\end{equation}
Intuitively, at early stages of the training phase, an imbalanced class distribution is used for representation learning of the feature extractor. 
At later stages, the model benefits more from a balanced dataset for training an unbiased classifier. 
Therefore, the control parameter $\alpha$ should progressively decrease from $1$ to $0$ during the training phase. Concretely, at training step $t$, $\alpha$ is calculated by~\cite{kang2019decoupling}:
$\alpha = 1 - \frac{t}{T}$, where $T$ is the total number of training epochs.

\subsection{Online Clustering Based PBS}
\label{sec:label}
The PBS method requires real labels to adjust the class distribution during training, which compromises its practicality. 
Since GCL is typically applied for the label-free scenario, we cannot directly apply PBS here. 
On one hand, \citep{mcpherson2001birds} have revealed the homophily phenomenon in homophilic graphs, \ie the nodes with similar features tend to be connected with each other and share the same label.
On the other hand, the motivation of GCL is to learn representations in which similar node pairs stay close to each other while dissimilar ones are far apart. 
We propose to connect these two facts by the pseudo-label method in GCL~\cite{caron2018deep}, which iteratively generates artificial labels by the model itself to make the PBS method applicable in the label-free GCL scenarios. 
We utilize the emergence of representation clusters learned from GCL to generate pseudo-labels of each node and then apply a down-sampling strategy to improve the quality of node representations in middle and tail classes.

Specifically, suppose there are $K$-classes for the node classification task.
At the certain iteration $t$ of the training phase, we obtain the node representation $\mZ_t \in \mathbb{R}^{N \times D}$ via the learned GNN encoder, where $D$ is the hidden dimension. We apply the clustering algorithm to the nodes in the embedding space to produce a set of $K$ prototypes $\{\vc_1,\dots,\vc_K\}$. Formally, we intend to learn a $D \times K$ centroid matrix $C$ and a one-hot cluster assignment vector $\hat{y}_n \in \mathbb{R}_{+}^{K}$ for each node $n$ of the given graph by solving the following problem:
\begin{equation}
    \min_{C \in \mathbb{R}^{D \times K}} \frac{1}{N} \sum_{n=1}^{N} \min_{\hat{y}_{n}}
    % \in\{0,1\}^{K}}
    \left\|\vz_{t,n}-C \hat{y}_{n}\right\|_{2}^{2} \; \text {such that} \quad \hat{y}_{n}^{\top} 1_{K}=1 ,
\end{equation}
where $\vz_{t,n} \in \mathbb{R}^D$ denotes the $n$-th node embedding vector of $\mZ_t$, and $1_{K}\in\mathbb{R}^K$ is the vector with all elements of $1$. Solving this above problem provides a set of optimal assignments $\{\hat{y}_n^*|n=1,\dots,N\}$ and a centroid matrix $C^*$. These assignments are then used as pseudo-labels.
Note that \emph{We set the number of centroids~(clusters) $K$ in the clustering algorithm equal to the number of classes in the training dataset}, which is an input hyperparameter of the classification task. We also perform the hyperparameter study in Appendix D and find that it is generally a good choice and prevents performance fluctuation. 
In order to avoid trivial solutions and empty clusters, we use the \emph{constrained K-means clustering}~\cite{bradley2000constrained} to instantiate the clustering algorithm. It can implement the K-means clustering algorithm whereby a minimum size for each cluster can be specified. Thus, we can address the representation collapse problem~\cite{fang2021exploring} which would produce a highly imbalanced pseudo-label distribution.

\paragraph{Theoretical Analysis.}
In order to justify the PBS method, we theoretically prove that, the classifier learned iteratively by balanced sampling with pseudo-labels on the imbalanced dataset can converge to the optimal balanced classifier with a linear rate. Detailed proofs can be found in Appendix B.

Consider a binary classification problem with two Gaussian distributions with different means and the equal variance. Suppose the data generating distribution is $P_{XY}$ and the probability of positive labels $(+1)$ and negative labels $(-1)$ are $P_Y(1)$ and $P_Y(-1)$, respectively. We have $X|Y=+1 \sim \mathcal{N}(\mu_1, \sigma^2)$ conditioned on $Y=+1$ and similarly, $X|Y=-1 \sim \mathcal{N}(\mu_2, \sigma^2)$ conditioned on $Y=-1$. In addition, suppose $\mu_1 < \mu_2$ without loss of generality. 
% We consider the imbalanced setting and 
It is straightforward to verify that~\cite{bishop2006pattern} the optimal decision boundary of a balanced Bayes classifier is
$\theta^* \equiv \frac{\mu_1+\mu_2}{2}$.
% Note that the training dataset is imbalanced but the testing dataset is balanced in imbalanced learning. However, in self-supervised GCL, label information is absent and we instead propose to create pseudo-labels iteratively via representations learned from GCL to enable the PBS method. 
At the first iteration $t=1$, we start from the imbalanced unlabeled dataset and generate pseudo-labels $\hat{Y}_0$ by the clustering method. Then, we obtain the estimated decision boundary $\hat{\theta}_1=\frac{\hat{\mu}_1 + \hat{\mu}_2}{2}$ and produce pseudo-labels $\hat{Y}_1$. 
The fact that the initial dataset being imbalanced ($P_Y(1)\neq P_Y(-1)$) leads to biased $\hat{\theta}_1$. 
We iteratively obtain the estimator $\hat{\theta}_t$ and pseudo-labels $\hat{Y}_t$ when $t \geq 2$.
% which has three consequences: (1) The learned classifier $\hat{f}_1$ exhibits higher recall/lower precision for the head classes and the opposite for the tail classes~\cite{yang2020rethinking}, aligned with our empirical observations in Fig.~\ref{fig:per-class}. (2) For any iteration $t$, the biased classifier $\hat{f}_t$ and pseudo-labels $\hat{Y}_t$ induce distribution shift between the pseudo-classes $\hat{X}$ and ground truth classes $X$. (3) For any following iteration $t$, the dataset re-balanced based on $\hat{Y}_t$ is still imbalanced with respect to the ground truth labels $Y$, which means $|\hat{\theta}_{t+1} - \theta^*| > 0, \forall t \in\mathbb{N}_{+}$. For which we prove the following proposition:

\begin{prop}
\label{prop}
Consider the above setup. Suppose there is a (sufficiently large) integer $T$ such that $|\hat{\theta}_T-\theta^*|\ll |\mu_2-\mu_1|$, $|(\hat{\theta}_T-\theta^*)(\mu_2-\mu_1)|\ll\sigma^2$.
Our estimator $\hat{\theta}_T$ converges to the optimal balanced decision boundary $\theta^*$, \ie $|\hat{\theta}_{t+1} - \theta^*| \leq C \cdot |\hat{\theta}_{t} - \theta^*|, \forall t\ge T$ with a linear convergence rate $C = \frac{2}{\pi} < 1$.
\end{prop}

\paragraph{Interpretation.} 
In the context of PBS, the classifier is trained starting from the imbalanced data distribution. Intuitively, by iteratively down-sampling head classes, the data distribution gradually becomes balanced, on which the trained classifier also converges to the balanced optimum.
% As stated in proposition~\ref{prop}, the classifier learned by balanced sampling with pseudo-label on the imbalanced dataset can converge to the optimal balanced classifier with a linear rate, which justifies the rationale of the PBS method.

\subsection{Node Centrality Based PBS}
\label{sec:ncpbs}
% To improve the representations of nodes in under-represented~(middle and tail) classes, we propose to down-sample nodes of the head classes to balance the class distribution over all classes.
To further improve the representations of nodes in under-represented~(middle and tail) classes, we incorporate graph structural information when performing the node centrality based PBS method.
% to achieve the full potential of the online clustering based PBS method.
In network science, node centrality, which measures how important a node is in a graph, is an important metric to understand the influence of each node of a graph~\cite{barabasi2013network}. 
Therefore, we propose an adaptive down-sampling scheme based on node centrality to balance the class distribution over all classes.
% when performing node centrality based PBS.
For each node class, we sample nodes of higher centrality with higher probability to better preserve the intrinsic structures of graphs in learned representations.
We herein utilize the PageRank centrality due to its simplicity and effectiveness~\cite{barabasi2013network}. 
Formally, the centrality values are calculated by the iterative form: $\bm{\sigma} = \alpha\mA\mD^{-1} + \bf{1}$, where $\bm{\sigma} \in \mathbb{R}^N$ is the PageRank centrality score vector for each node, $\alpha$ is a damping factor to control the probability of randomly jumping to another node in the graph, $A$ and $D$ denote the adjacency and the degree matrix of the input graph respectively, and $\bf{1}$ is the all-ones identity vector. 
Note that we pre-calculate the PageRank score $\sigma$ before the training phase of GCL. When performing down-sampling of the head classes, we calculate the probability of each node based on $\sigma$. 
Formally, for node $v$ in class $j$ with the centrality score $\sigma_v$, the probability with the normalized centrality score is defined as: 
\begin{equation}
    p^{\text{NPB}}_{v,j} = \max \left\{ \frac{\bm{\sigma}_{v} - \bm{\sigma}_{\min}}{\bm{\sigma}_{\max}-\bm{\sigma}_{\min}} \cdot p^{\text{PB}}_{j}, p_{\tau} \right\},
\label{eq:5}
\end{equation}
where $p^{\text{PB}}_{j}$ is the progressively balanced sampling probability of class $j$, $\bm{\sigma}_{\max}$ and $\bm{\sigma}_{\min}$ are the maximum and minimum value of the centrality score, and $p_{\tau}$ is a cut-off probability to ensure that nodes with extremely low probabilities can also be sampled. 
In node centrality based PBS, we then perform a normalization step that transforms $p^{\text{NPB}}_{v,j}$ into probabilities and then use it to balance the class distribution. 
Concretely, we select certain nodes of the original graph in form of a \emph{masked vector} $\vm \in \mathbb{R}^N$ by sampling each node independently according to $p^{\text{NPB}}_{v,j}$. We then calculate the graph contrastive loss only on these selected node representations, as shown in Fig.~\ref{fig:framework}.

\begin{figure*}[h]
    \centering
    \includegraphics[width=.78\linewidth]{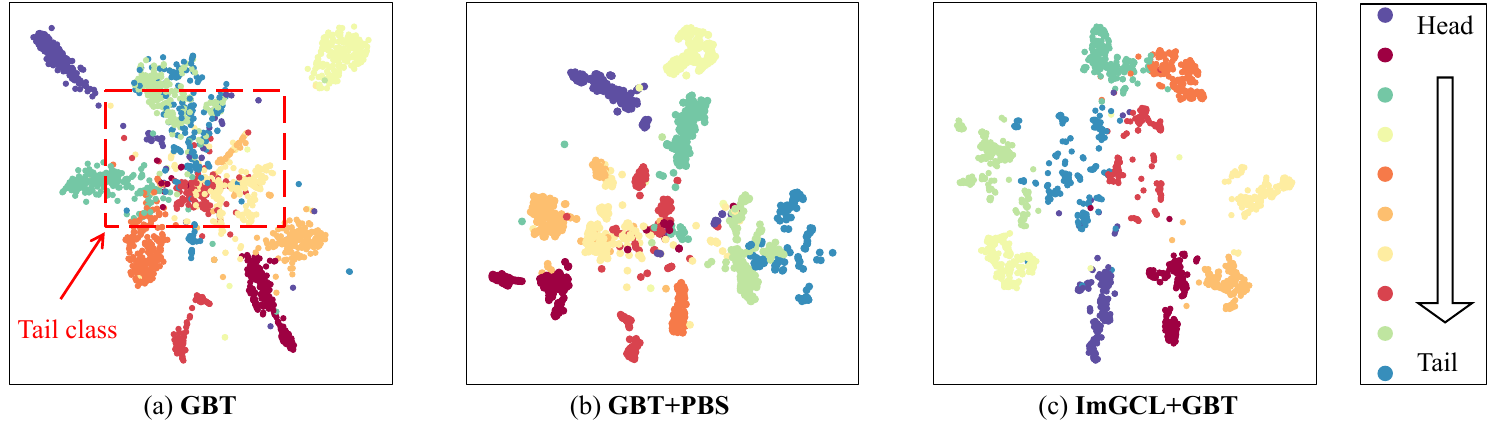}
    \vspace{-0.2em}
    \caption{Visualization of the testing set on Amazon-Computers. Each point in the figure is colored by real labels.}
    \vspace{-0.3em}
    \label{fig:tsne}
\end{figure*}

\subsection{Learning Framework}
ImGCL is a general GCL framework for imbalanced node classification, which can be readily applied with existing GCL methods adopting the two-branch design~\cite{thakoor2021bootstrapped, bielak2021graph, you2020graph}. ImGCL does not rely on specific approaches of graph view augmentation, graph view encoding, or representation contrasting in GCL. In ImGCL, we set the number of clusters $K$ in the node centrality based PBS method equal to the number of classes in the downstream task.
One challenge of the implementation is that the quality of the generated pseudo-labels and representations are mutually-dependent, which could destabilize the training loop. In response, we re-balance the class distribution every $B$ epochs, thus there are $T / B$ times to adjust the class distribution. 
In order to improve the representations of nodes in under-represented~(middle and tail) classes, we select $N\times l$ nodes during the pre-training phase in ImGCL, where $l=10\%$ equals the ratio of training data in the down-stream task.
We down-sample the head nodes to the required number according to their PageRank centrality scores as introduced in Sec.~\ref{sec:ncpbs}.
Following the linear evaluation scheme of GCL~\cite{zhu2020graph}, we train a linear classifier on the balanced dataset with 10\% randomly selected nodes after obtaining node representations.
The training algorithm of ImGCL is summarized as follows.

\begin{algorithm}[h]
    \caption{The ImGCL pre-training algorithm}
    \label{alg}
    \KwIn{The input graph $\gG$, GNN encoder $\mathcal{F}$.}
    \Parameter{Number of nodes $N$, number of clusters $K$, re-balanced labeling frequency $B$, the ratio of selected nodes $l$.}
  	\KwOut{Pre-trained GNN encoder $\mathcal{F}$.}
  	Calculate the node centrality vector $\sigma$ of $\gG$. \\
    \For{$epoch = 0, 1, 2, \cdots$}{
        Draw two augmentation functions $t_1 \sim \mathcal{T}$, $t_2 \sim \mathcal{T}$. \\
        Generate two graph views $\widetilde{\gG}_1=t_1(\gG)$ and $\widetilde{\gG}_2=t_2(\gG)$. \\
        Obtain node representations $U$ of $\widetilde{\gG}_1$ and $V$ of $\widetilde{\gG}_2$ using the GNN encoder $\mathcal{F}$. \\
        \If{$epoch \; mod \; B == 0$}{
            Cluster node representations to obtain pseudo-labels.\\
            Calculate the normalized centrality score $p^{\text{NPB}}$ with Eq.~\ref{eq:5}. \\
            Obtain the masked vector $m$ according to $p^{\text{NPB}}$, which satisfies $\|m\| = N \times l$. \\
        }
        Compute the contrastive object $\mathcal{L}$ on these selected node representations $U \odot m$ and $V \odot m$ with Eq.~\ref{equ:1}. \\
        Update the parameters of $\mathcal{F}$ with $\mathcal{L}$.
    }
\end{algorithm} 
\section{Case Study: Learning Discriminative Representations on Amazon-Computers}
We are particularly interested in the learned representations among different methods.
For a more intuitive comparison and to further demonstrate the effectiveness of the ImGCL framework, we design experiments of visualization on Amazon-computers. We utilize the output representations on the last layer of vanilla GBT, GBT with PBS to train the classifier on training data but without node centrality based PBS on representation learning in GCL, and GBT with ImGCL. We plot the learned representations of the testing graph dataset using t-SNE~\cite{van2008visualizing}. 
As shown in Fig.~\ref{fig:tsne}, vanilla GBT clearly exhibits the minority collapse~\cite{fang2021exploring} phenomenon, and the representations in tail classes are mixed together, which fundamentally limits the performance in the tail classes. 
In Fig.~\ref{fig:tsne} (b), we can find clearer boundaries among different classes but the representations in tail classes are still mixed together, which suggests that it is important to explore balanced representation spaces in GCL methods.
In comparison, the learned representations of GBT+ImGCL~(Fig.~\ref{fig:tsne} (c)) have distinct boundaries among different classes and more compact intra-class structures, which highlights the effectiveness of our proposed ImGCL framework.
\section{Experiments}
\label{sec:experiments}
% ~\footnote{Our codes and data are publicly available at \url{https://tinyurl.com/ImGCL}.}:
In this section, we provide empirical results to demonstrate the effectiveness of our ImGCL framework. We conduct extensive experiments on imbalanced graph datasets to mainly answer the following questions:~\footnote{Due to space limitations, ablations on different components in ImGCL and hyperparameter studies are provided in the appendix.}
(1) Can ImGCL generally improve the performance of GCL methods on imbalanced node classification?~(Sec.~\ref{sec:exp_node})
(2) How does ImGCL perform on different imbalanced types?~(Sec.~\ref{sec:exp_analysis})
(3) How does ImGCL help improve GCL methods on imbalanced node classification?~(Sec.~\ref{sec:exp_analysis})

% \usepackage{multirow}
\begin{table*}[]
\centering
\caption{Summary of accuracies~(\%) with standard deviation on imbalanced node classification. The 'Available Data' means data we can obtain for training, where $\mX$, $\mA$, and $\mY$ denote node features, the adjacency matrix, and labels respectively. We highlight models in the ImGCL category with the gray background. The highest performance under each category is masked as \textbf{bold}. The highest performance improvement of the GCL baseline w $\&$ w/o the ImGCL framework is \underline{underlined}.}
% \vspace{-1em}
\label{table:main}
\setlength{\tabcolsep}{9pt}
\scalebox{0.83}{
\begin{tabular}{ccccccc}
\toprule
Category                   & Method            & Available Data & Amazon-Computers          & Amazon-Photo              & Wiki-CS                   & DBLP                      \\ \midrule
\multirow{4}{*}{}        & Raw features      & \multicolumn{1}{c|}{\textbf{\emph{X}}}              & $33.99_{\pm4.47}$               & $38.07_{\pm4.03}$               & $34.50_{\pm2.06}$               & $38.51_{\pm0.80}$               \\
                            & Node2vec          & \multicolumn{1}{c|}{\textbf{\emph{A}}}              & $69.80_{\pm2.69}$               & $69.44_{\pm0.38}$               & $\bm{51.76_{\pm2.24}}$               & $50.41_{\pm2.77}$               \\
                            & DeepWalk          & \multicolumn{1}{c|}{\textbf{\emph{A}}}              & $69.67_{\pm2.36}$               & $69.00_{\pm0.62}$               & $51.32_{\pm2.17}$               & $\bm{50.57_{\pm2.88}}$               \\
                            & DeepWalk + features & \multicolumn{1}{c|}{\textbf{\emph{X, A}}}           &   $\bm{70.20_{\pm3.30}}$    &                 $\bm{71.60_{\pm3.31}}$   &                 $51.51_{\pm2.51}$          &          $49.57_{\pm1.65}$                                        \\ \midrule
\multirow{6}{*}{GCL}        & DGI               & \multicolumn{1}{c|}{\textbf{\emph{X, A}}}           & $10.88_{\pm2.09}$               & $\underline{13.62_{\pm2.26}}$               & $17.11_{\pm4.83}$               & $\underline{26.63_{\pm10.87}}$              \\
                            & MVGRL             & \multicolumn{1}{c|}{\textbf{\emph{X, A}}}           & $13.40_{\pm3.01}$               & $16.92_{\pm3.14}$               & $45.97_{\pm2.42}$               & $44.43_{\pm0.57}$               \\
                            & InfoGraph         & \multicolumn{1}{c|}{\textbf{\emph{X, A}}}           & $\underline{35.83_{\pm7.01}}$               & $53.57_{\pm13.29}$              & $44.19_{\pm3.90}$               & $48.26_{\pm5.95}$               \\
                            & GRACE             & \multicolumn{1}{c|}{\textbf{\emph{X, A}}}           & $\bm{41.54_{\pm2.51}}$               & $45.24_{\pm4.24}$               & $\bm{54.20_{\pm3.97}}$               & $44.48_{\pm0.40}$               \\
                            & BGRL              & \multicolumn{1}{c|}{\textbf{\emph{X, A}}}           & $40.81_{\pm5.01}$               & $51.18_{\pm10.49}$              & $39.82_{\pm3.60}$               & $49.58_{\pm3.99}$               \\
                            & GBT               & \multicolumn{1}{c|}{\textbf{\emph{X, A}}}           & $42.17_{\pm3.74}$               & $\bm{60.73_{\pm3.64}}$               & $\underline{44.95_{\pm2.63}}$               & $\bm{58.51_{\pm5.30}}$               \\ \midrule
\rowcolor{black!10}   & DGI               & \multicolumn{1}{c|}{\textbf{\emph{X, A}}}           & $48.85_{\pm10.94}$              & $\underline{47.99_{\pm9.03}}$               & $41.20_{\pm18.84}$              & $\underline{50.39_{\pm9.17}}$               \\
                            \rowcolor{black!10}& MVGRL             & \multicolumn{1}{c|}{\textbf{\emph{X, A}}}           & $46.42_{\pm11.33}$              & $50.86_{\pm9.49}$               & $60.85_{\pm2.60}$               & $51.90_{\pm7.42}$               \\
                            \rowcolor{black!10}& InfoGraph         & \multicolumn{1}{c|}{\textbf{\emph{X, A}}}           & $\underline{75.44_{\pm5.30}}$               & $72.56_{\pm2.91}$               & $68.96_{\pm5.86}$               & $69.12_{\pm3.24}$               \\
                            \rowcolor{black!10}ImGCL (ours) & GRACE             & \multicolumn{1}{c|}{\textbf{\emph{X, A}}}           & $77.54_{\pm3.00}$               & $68.89_{\pm4.41}$               & $\bm{73.86_{\pm2.78}}$               & $63.61_{\pm4.91}$               \\
                            \rowcolor{black!10}& BGRL              & \multicolumn{1}{c|}{\textbf{\emph{X, A}}}           & $67.82_{\pm10.24}$              & $72.67_{\pm6.04}$               & $59.35_{\pm15.44}$              & $57.90_{\pm2.28}$               \\
                            \rowcolor{black!10}& GBT               & \multicolumn{1}{c|}{\textbf{\emph{X, A}}}           & $\bm{78.62_{\pm1.73}}$               & $\bm{75.13_{\pm7.13}}$               & $\underline{73.08_{\pm12.45}}$              & $\bm{70.05_{\pm1.78}}$               \\ \midrule
\multicolumn{3}{c|}{\textbf{Best ImGCL over GCL}}                 & \multicolumn{1}{c}{\color{red}\textbf{39.61  $\uparrow$}} & \multicolumn{1}{c}{\color{red}\textbf{34.47  $\uparrow$}} & \multicolumn{1}{c}{\color{red}\textbf{28.13  $\uparrow$}} & \multicolumn{1}{c}{\color{red}\textbf{23.76  $\uparrow$}} \\ \midrule
\multirow{2}{*}{Supervised} & GCN               & \multicolumn{1}{c|}{\textbf{\emph{X, A, Y}}}        & $46.83_{\pm1.52}$               & $68.84_{\pm2.56}$               & $59.84_{\pm2.02}$               & $51.55_{\pm2.64}$               \\
                            & GCN+PBS              & \multicolumn{1}{c|}{\textbf{\emph{X, A, Y}}}        & $\bm{70.12_{\pm9.78}}$               & $\bm{73.34_{\pm8.28}}$               & $\bm{63.15_{\pm5.13}}$               & $\bm{73.11_{\pm2.76}}$               \\ \bottomrule
\end{tabular}
}
\vspace{-1em}
\end{table*}

\label{sec:exp_setup}
\paragraph{Dataset.}
We use four widely-used datasets including Wiki-CS, Amazon-computers, Amazon-photo, and DBLP, to comprehensively study the performance of transductive node classification. 
Recently proposed works evaluate the performance on semi-supervised node classification.
ImGAGN~\cite{qu2021imgagn} reconstructs the binary imbalanced network by setting the smallest class as the tail class and the rest classes as the head class. The testing dataset in Tail-GNN~\cite{liu2021tail} is also imbalanced. We think these experimental settings are not ideal for imbalanced learning due to the limited number of node classes or imbalanced testing datasets.
In order to validate the effectiveness of ImGCL on the imbalanced node classification setting, 
we select an equal number of nodes in each class for the validation and testing dataset. Following~\cite{zhu2020graph}, the training set is randomly sampled from the rest according to train/valid/test ratios = 1:1:8, which is highly imbalanced.
The descriptions, statistics, and the imbalance ratio of each dataset can be found in Appendix H. 
% A detailed description of datasets can be found in Appendix.

\paragraph{Evaluation Protocol.} 
For each experiment, we follow the commonly-used \emph{linear evaluation scheme} for GCL as introduced in~\cite{zhu2020graph}. The model is firstly trained in a self-supervised manner, and then the learned representations are used to train and test with a simple linear classiﬁer. For results in this section, we train each model in twenty runs for different data splits and report the average performance with the corresponding standard deviation for a fair comparison. In what follows, we measure performance in terms of accuracy, if not otherwise specified.

\paragraph{Imbalanced Types.}
In order to comprehensively evaluate the performance of ImGCL in different imbalanced types, we introduce two imbalanced types~\cite{jiang2021self}: \emph{Exp} and \emph{Pareto}, parameterized by an imbalanced factor. \emph{Exp} imbalanced class distribution is given by an exponential function, where the higher imbalanced factor means the more imbalanced graph. \emph{Pareto} imbalanced class distribution is determined by a Pareto distribution, where a lower imbalanced factor means the smaller power value of a Pareto distribution and thus the more imbalanced graph. 
% Pareto distribution with imbalanced factor $1$ means that nodes of a graph follow a power-law probability distribution,  where $90\%$ of nodes are from the first $10\%$ classes.

\paragraph{Baselines.} We consider representative baseline methods in the following two categories: (1) traditional methods including Node2vec~\cite{grover2016node2vec}, DeepWalk~\cite{perozzi2014deepwalk}, and raw features as input without considering the graph topology.
(2) deep learning methods including DGI~\cite{velivckovic2018deep}, MVGRL~\cite{hassani2020contrastive}, InfoGraph~\cite{sun2019infograph}, GRACE~\cite{zhu2020graph}, BGRL~\cite{thakoor2021bootstrapped}, and GBT~\cite{bielak2021graph}. 
We also directly compare ImGCL with the supervised counterparts, \ie the most representative model GCN~\cite{kipf2016semi} and the variant of GCN trained with PBS.
Note that for all baselines, we report their performance on the imbalanced experimental settings following their official hyperparameters~(detailed in Appendix H) based on the PyGCL~\cite{zhu2021empirical} open-source library.

\subsection{Experimental Results on Node Classification}
\label{sec:exp_node}
The empirical performance of imbalanced node classification with the \emph{Exp} type and $100$ imbalanced factor is summarized in Table~\ref{table:main}. ImGCL consistently outperforms current GCL baselines or even the supervised baselines. We summarize our observations from the table as follows: (1) 
Recently proposed GCL methods \cite{zhu2021empirical}, which are evaluated on balanced testing sets, exhibit severe performance degradation in our imbalanced node classification setting.
By incorporating our proposed ImGCL framework~(the gray background), these GCL methods improve by a large margin. Concretely, ImGCL+GCL achieves [$39.61\%$, $34.47\%$, $28.13\%$, $23.76\%$] average absolute gain in accuracy than the baseline GCL models on [Amazon-Computers, Amazon-Photo, Wiki-CS, DBLP], respectively. We also find that the recently proposed GBT~\cite{bielak2021graph} obtains the best performance among a set of GCL competitors. We think the reason is that \emph{feature decorrelation} method in GBT is more fit for the imbalanced node classification and we adopt the GBT baseline model in the following experimental analysis. 
(2) The traditional methods, \eg Node2vec and DeepWalk, can achieve competitive performance in the imbalanced node classification task compared with GCL methods. We postulate the reason is that the traditional network embedding methods can take advantage of the homophily~\cite{barabasi2013network} property to utilize the graph topology features which is important on imbalanced node classification, as reflected in~\cite{liu2021tail}. Moreover, the ``Raw features'' method without considering the graph topology cannot achieve satisfactory performance, which indicates the necessity of utilizing the graph topology features on imbalanced node classification.
(3) Compared with the supervised learning methods GCN and GCN+PBS, the ImGCL framework achieves superior or competitive performance on all datasets, which further corroborates the effectiveness of our proposed framework.

\begin{table}[t]
\centering\caption{Results of accuracy~(\%) on Amazon-computers using the GBT baseline model under different imbalanced types and factors. $\uparrow$ means a higher imbalanced factor corresponds to a more imbalanced dataset. Instead, $\downarrow$ means a lower factor corresponds to a more imbalanced dataset. All models are trained for eight iterations of pseudo labeling.}
\label{table:imbalanced-type}
\setlength{\tabcolsep}{3pt}
\vspace{-0.5em}
\scalebox{0.67}{
\begin{tabular}{ccccccc}
\toprule
{\begin{tabular}{c}Type\end{tabular}}         & {\begin{tabular}{c}Factor\end{tabular}} & {\begin{tabular}{c}Method\\Catergory\end{tabular}}                & Head         & Middle & Tail          & All         \\ \midrule
\multirow{8}{*}{Exp\;$\uparrow$}    & \multicolumn{1}{|c|}{20}               & \multirow{4}{*}{GBT}       & \multicolumn{1}{|c}{$79.45_{\pm3.59}$}  & $65.40_{\pm6.37}$  & $73.25_{\pm4.52}$  & $71.97_{\pm3.78}$ \\
                        & \multicolumn{1}{|c|}{50}               &                                 & \multicolumn{1}{|c}{$80.12_{\pm2.91}$}  & $60.57_{\pm3.35}$  & $57.42_{\pm7.92}$  & $65.49_{\pm3.37}$ \\
                        & \multicolumn{1}{|c|}{100}              &                                 & \multicolumn{1}{|c}{$70.38_{\pm5.96}$}  & $41.92_{\pm3.28}$  & $14.29_{\pm3.53}$    & $42.17_{\pm3.74}$ \\
                        & \multicolumn{1}{|c|}{200}              &                                 & \multicolumn{1}{|c}{$78.13_{\pm2.56}$}  & $25.12_{\pm4.61}$  & $10.28_{\pm3.28}$   & $36.57_{\pm4.30}$ \\\cmidrule{2-7}
                        \rowcolor{black!10}& \multicolumn{1}{|c|}{20}               &  & \multicolumn{1}{|c}{$71.58_{\pm7.77}$}  & $89.95_{\pm5.20}$  & $94.55_{\pm3.36}$  & $85.82_{\pm1.99}$ \\
                        \rowcolor{black!10}& \multicolumn{1}{|c|}{50}               &                                & \multicolumn{1}{|c}{$63.88_{\pm11.73}$} & $86.03_{\pm9.92}$  & $95.75_{\pm3.83}$  & $82.30_{\pm3.69}$ \\
                        \rowcolor{black!10}& \multicolumn{1}{|c|}{100}              &                               & \multicolumn{1}{|c}{$57.32_{\pm7.29}$}  & $82.72_{\pm8.46}$  & $94.45_{\pm6.07}$  & $78.62_{\pm1.73}$ \\
                        \rowcolor{black!10}& \multicolumn{1}{|c|}{200}              &          \multirow{-4}{*}{{\begin{tabular}{c}GBT+ImGCL\\(ours)\end{tabular}}}                    & \multicolumn{1}{|c}{$46.95_{\pm17.70}$} & $75.27_{\pm18.75}$ & $76.42_{\pm18.19}$ & $67.12_{\pm8.14}$ \\ \midrule\midrule
\multirow{4}{*}{Pareto\;$\downarrow$} & \multicolumn{1}{|c|}{2}                & \multirow{2}{*}{GBT}       & \multicolumn{1}{|c}{$83.31_{\pm1.42}$}  & $61.01_{\pm1.07}$  & $52.58_{\pm9.05}$  & $65.17_{\pm1.66}$ \\
                        & \multicolumn{1}{|c|}{1}                &                                 & \multicolumn{1}{|c}{$82.06_{\pm1.44}$}  & $57.22_{\pm2.01}$  & $39.85_{\pm7.25}$  & $59.46_{\pm1.74}$ \\\cmidrule{2-7}
                        \rowcolor{black!10}& \multicolumn{1}{|c|}{2}                &  & \multicolumn{1}{|c}{$53.81_{\pm3.19}$}  & $70.25_{\pm2.41}$  & $93.19_{\pm2.13}$  & $72.20_{\pm0.78}$ \\
                        \rowcolor{black!10}& \multicolumn{1}{|c|}{1}                &           \multirow{-2}{*}{{\begin{tabular}{c}GBT+ImGCL\\(ours)\end{tabular}}}                      & \multicolumn{1}{|c}{$41.94_{\pm4.75}$}  & $78.34_{\pm2.84}$  & $87.97_{\pm3.06}$  & $70.31_{\pm1.38}$ \\ \bottomrule
\end{tabular}
}
\vspace{-1em}
\end{table}

\subsection{Experimental Analysis}
\label{sec:exp_analysis}
\paragraph{Imbalanced Type Analysis.}
The performance for the \emph{head, middle, and tail} classes are usually reported on imbalanced learning~\cite{zhang2021deep}.
We first divide the training set of Amazon-computers into three disjoint groups in terms of class size: \{\emph{Head}, \emph{Middle}, \emph{Tail}\}. \emph{Head} and \emph{Tail} each include the top and bottom $1/3$ classes, respectively. Because there are 10 classes of Amazon-computers in total, the classes with sorted indices in decreasing order [1-3, 4-7, 8-10] belong to [\emph{Head}~(3 classes), \emph{Middle}~(4 classes), \emph{Tail}~(3 classes)] categories, respectively.
To validate ImGCL across different imbalanced class distributions, we design experiments on Amazon-computers using the GBT baseline model w $\&$ w/o the ImGCL framework.
We consider two imbalanced types: Exp and Pareto. Four imbalanced factors~[20, 50, 100, 200] are associated with Exp.
Two imbalance factors~[1, 2] are chosen for Pareto.
In Table~\ref{table:imbalanced-type}, we observe that the more imbalanced dataset leads to the lower accuracy of the model. Intuitively, a too imbalanced dataset might break our assumption in Proposition~\ref{prop} that $|\hat{\theta}_t-\theta^*|\ll |\mu_2-\mu_1|$, which undermines the convergence guarantee and therefore degrades the performance of ImGCL given a fixed number of iterations. Nevertheless, the ImGCL+baseline model consistently outperforms the baseline model.

\begin{figure}[t]
    \includegraphics[width=1.\linewidth]{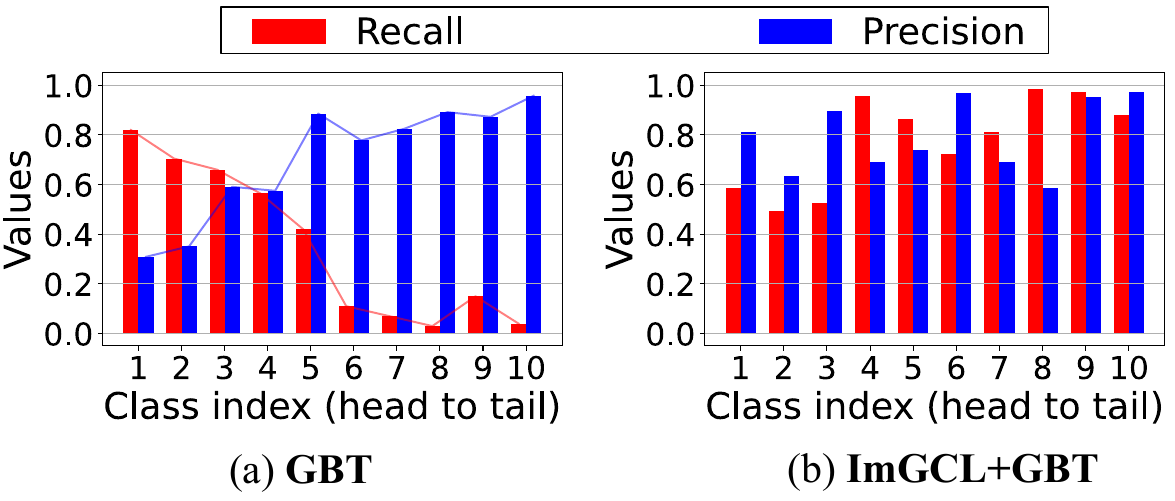}
    \vspace{-1.8em}
    \caption{Bias comparison between GBT models w \& w/o ImGCL on Amazon-Computers under imbalanced experimental settings. \textbf{Left:} Per-class recall and precision w/o ImGCL. \textbf{Right:} Per-class recall and precision with ImGCL. The class index is sorted by the number of nodes in each class in descending order. GBT w/o ImGCL clearly shows a descending trend in recall while an ascending trend in precision. However, GBT with ImGCL has achieved a relatively balanced recall and precision value over all classes.} 
    \label{fig:per-class}
    \vspace{-1.2em}
\end{figure}
\paragraph{Per-Class Analysis.}
\label{sec:exp_perclass}
We analyze the per-class recall and precision in Fig.~\ref{fig:per-class} to better understand how ImGCL can help improve GCL methods on the imbalanced node classification. We test GBT on Amazon-computers under imbalanced experimental settings introduced in Sec.~\ref{sec:exp_setup}. The GBT baseline model exhibits highly skewed performance on head and tail classes. The recall on the most majority and minority class is 82.0\% and 4.1\% respectively, while the corresponding numbers for precision are 30.9\% and 95.8\%.
We observe that GBT falsely classifies most of the tail class samples into head classes with low conﬁdence but have high conﬁdence on the correctly classiﬁed nodes in tail classes~\cite{wei2021crest}. In contrast, GBT with ImGCL obtains relatively balanced recall and precision value on both head (the most majority precision: 30.9\% $\rightarrow$ 81.2\%) and tail (the most minority recall: 4.1\% $\rightarrow$ 88.2\%) classes, which leads to the substantial improvement in the overall accuracy (\ie 42.17\% $\rightarrow$ 78.62\%) across all classes, as shown in Table 3.
\section{Conclusion}
\label{sec:conclusion}
In this work, we study how to improve the representations of graph contrastive learning~(GCL) methods on imbalanced node classification, which is a very practical but rarely explored problem. We propose the principled ImGCL framework, which automatically and adaptively balances the representations learned from GCL without knowing labels and then theoretically justifies it. Through extensive experiments on multiple graph datasets and imbalance settings, we show that ImGCL can significantly improve the recently proposed GCL methods by improving the representations of nodes in under-represented~(tail) classes.
For the future work, we will explore more data types, such as bioinformatics graphs. We hope our work will extend GCL to more realistic task settings with (underlying) imbalanced node class distribution.

% \clearpage
% \bibliographystyle{unsrtnat}
\section*{Acknowledgments}
Liang Zeng and Jian Li are supported in part by the National Natural Science Foundation of China Grant 62161146004, Turing AI Institute of Nanjing and Xi'an Institute for Interdisciplinary Information Core Technology.
\bibliography{aaai23}

\begin{thebibliography}{45}
\providecommand{\natexlab}[1]{#1}

\bibitem[{Barab{\'a}si(2013)}]{barabasi2013network}
Barab{\'a}si, A.-L. 2013.
\newblock Network science.
\newblock \emph{Philosophical Transactions of the Royal Society A:
  Mathematical, Physical and Engineering Sciences}, 371(1987): 20120375.

\bibitem[{Barab{\'a}si and Albert(1999)}]{barabasi1999emergence}
Barab{\'a}si, A.-L.; and Albert, R. 1999.
\newblock Emergence of scaling in random networks.
\newblock \emph{science}, 286(5439): 509--512.

\bibitem[{Bielak, Kajdanowicz, and Chawla(2021)}]{bielak2021graph}
Bielak, P.; Kajdanowicz, T.; and Chawla, N.~V. 2021.
\newblock Graph Barlow Twins: A self-supervised representation learning
  framework for graphs.
\newblock \emph{arXiv preprint arXiv:2106.02466}.

\bibitem[{Bishop and Nasrabadi(2006)}]{bishop2006pattern}
Bishop, C.~M.; and Nasrabadi, N.~M. 2006.
\newblock \emph{Pattern recognition and machine learning}, volume~4.
\newblock Springer.

\bibitem[{Bradley, Bennett, and Demiriz(2000)}]{bradley2000constrained}
Bradley, P.~S.; Bennett, K.~P.; and Demiriz, A. 2000.
\newblock Constrained k-means clustering.
\newblock \emph{Microsoft Research, Redmond}, 20(0): 0.

\bibitem[{Caron et~al.(2018)Caron, Bojanowski, Joulin, and
  Douze}]{caron2018deep}
Caron, M.; Bojanowski, P.; Joulin, A.; and Douze, M. 2018.
\newblock Deep clustering for unsupervised learning of visual features.
\newblock In \emph{Proceedings of the European Conference on Computer Vision
  (ECCV)}, 132--149.

\bibitem[{Chawla et~al.(2002)Chawla, Bowyer, Hall, and
  Kegelmeyer}]{chawla2002smote}
Chawla, N.~V.; Bowyer, K.~W.; Hall, L.~O.; and Kegelmeyer, W.~P. 2002.
\newblock SMOTE: synthetic minority over-sampling technique.
\newblock \emph{Journal of artificial intelligence research}, 16: 321--357.

\bibitem[{Chen et~al.(2020)Chen, Kornblith, Norouzi, and
  Hinton}]{chen2020simpleframework}
Chen, T.; Kornblith, S.; Norouzi, M.; and Hinton, G. 2020.
\newblock A simple framework for contrastive learning of visual
  representations.
\newblock In \emph{International conference on machine learning (ICML)},
  1597--1607. PMLR.

\bibitem[{Fang et~al.(2021)Fang, He, Long, and Su}]{fang2021exploring}
Fang, C.; He, H.; Long, Q.; and Su, W.~J. 2021.
\newblock Exploring deep neural networks via layer-peeled model: Minority
  collapse in imbalanced training.
\newblock \emph{Proceedings of the National Academy of Sciences}, 118(43).

\bibitem[{Grover and Leskovec(2016)}]{grover2016node2vec}
Grover, A.; and Leskovec, J. 2016.
\newblock node2vec: Scalable feature learning for networks.
\newblock In \emph{Proceedings of the 22nd ACM SIGKDD international conference
  on Knowledge discovery and data mining (KDD)}, 855--864.

\bibitem[{Hamilton(2020)}]{hamilton2020graph}
Hamilton, W.~L. 2020.
\newblock Graph representation learning.
\newblock \emph{Synthesis Lectures on Artifical Intelligence and Machine
  Learning}, 14(3): 1--159.

\bibitem[{Hassani and Khasahmadi(2020)}]{hassani2020contrastive}
Hassani, K.; and Khasahmadi, A.~H. 2020.
\newblock Contrastive multi-view representation learning on graphs.
\newblock In \emph{International Conference on Machine Learning (ICML)},
  4116--4126. PMLR.

\bibitem[{He et~al.(2021)He, Kortylewski, Yang, Liu, Yang, Wang, and
  Yuille}]{he2021rethinking}
He, J.; Kortylewski, A.; Yang, S.; Liu, S.; Yang, C.; Wang, C.; and Yuille, A.
  2021.
\newblock Rethinking Re-Sampling in Imbalanced Semi-Supervised Learning.
\newblock \emph{arXiv preprint arXiv:2106.00209}.

\bibitem[{Jiang et~al.(2021)Jiang, Chen, Mortazavi, and Wang}]{jiang2021self}
Jiang, Z.; Chen, T.; Mortazavi, B.; and Wang, Z. 2021.
\newblock Self-Damaging Contrastive Learning.
\newblock In \emph{International Conference on Machine Learning (ICML)}.

\bibitem[{Kang et~al.(2020{\natexlab{a}})Kang, Li, Xie, Yuan, and
  Feng}]{kang2020exploring}
Kang, B.; Li, Y.; Xie, S.; Yuan, Z.; and Feng, J. 2020{\natexlab{a}}.
\newblock Exploring balanced feature spaces for representation learning.
\newblock In \emph{International Conference on Learning Representations
  (ICLR)}.

\bibitem[{Kang et~al.(2020{\natexlab{b}})Kang, Xie, Rohrbach, Yan, Gordo, Feng,
  and Kalantidis}]{kang2019decoupling}
Kang, B.; Xie, S.; Rohrbach, M.; Yan, Z.; Gordo, A.; Feng, J.; and Kalantidis,
  Y. 2020{\natexlab{b}}.
\newblock Decoupling representation and classifier for long-tailed recognition.
\newblock In \emph{8th International Conference on Learning Representations
  (ICLR)}.

\bibitem[{Kipf and Welling(2016)}]{kipf2016semi}
Kipf, T.~N.; and Welling, M. 2016.
\newblock Semi-supervised classification with graph convolutional networks.
\newblock In \emph{Proceedings of the International Conference on Learning
  Representations (ICLR)}.

\bibitem[{Kumar et~al.(2018)Kumar, Hooi, Makhija, Kumar, Faloutsos, and
  Subrahmanian}]{kumar2018rev2}
Kumar, S.; Hooi, B.; Makhija, D.; Kumar, M.; Faloutsos, C.; and Subrahmanian,
  V. 2018.
\newblock Rev2: Fraudulent user prediction in rating platforms.
\newblock In \emph{Proceedings of the Eleventh ACM International Conference on
  Web Search and Data Mining (WSDM)}, 333--341.

\bibitem[{Li et~al.(2022)Li, Zeng, Gao, Yuan, Bian, Wu, Zhang, Lu, Yu, Liu
  et~al.}]{li2022imdrug}
Li, L.; Zeng, L.; Gao, Z.; Yuan, S.; Bian, Y.; Wu, B.; Zhang, H.; Lu, C.; Yu,
  Y.; Liu, W.; et~al. 2022.
\newblock ImDrug: A Benchmark for Deep Imbalanced Learning in AI-aided Drug
  Discovery.
\newblock \emph{arXiv preprint arXiv:2209.07921}.

\bibitem[{Liu et~al.(2019)Liu, Miao, Zhan, Wang, Gong, and Yu}]{liu2019large}
Liu, Z.; Miao, Z.; Zhan, X.; Wang, J.; Gong, B.; and Yu, S.~X. 2019.
\newblock Large-scale long-tailed recognition in an open world.
\newblock In \emph{Proceedings of the IEEE/CVF Conference on Computer Vision
  and Pattern Recognition (CVPR)}, 2537--2546.

\bibitem[{Liu, Nguyen, and Fang(2021)}]{liu2021tail}
Liu, Z.; Nguyen, T.-K.; and Fang, Y. 2021.
\newblock Tail-GNN: Tail-Node Graph Neural Networks.
\newblock In \emph{Proceedings of the 27th ACM SIGKDD Conference on Knowledge
  Discovery and Data Mining (KDD)}, 1109--1119.

\bibitem[{Liu et~al.(2020)Liu, Zhang, Fang, Zhang, and Hoi}]{liu2020towards}
Liu, Z.; Zhang, W.; Fang, Y.; Zhang, X.; and Hoi, S.~C. 2020.
\newblock Towards locality-aware meta-learning of tail node embeddings on
  networks.
\newblock In \emph{Proceedings of the 29th ACM International Conference on
  Information and Knowledge Management (CIKM)}, 975--984.

\bibitem[{Ma et~al.(2020)Ma, Bian, Rong, Huang, Xu, Xie, Ye, and
  Huang}]{ma2020multi}
Ma, H.; Bian, Y.; Rong, Y.; Huang, W.; Xu, T.; Xie, W.; Ye, G.; and Huang, J.
  2020.
\newblock Multi-View Graph Neural Networks for Molecular Property Prediction.
\newblock \emph{arXiv preprint arXiv:2005.13607}.

\bibitem[{McPherson, Smith-Lovin, and Cook(2001)}]{mcpherson2001birds}
McPherson, M.; Smith-Lovin, L.; and Cook, J.~M. 2001.
\newblock Birds of a feather: Homophily in social networks.
\newblock \emph{Annual review of sociology}, 27(1): 415--444.

\bibitem[{Perozzi, Al-Rfou, and Skiena(2014)}]{perozzi2014deepwalk}
Perozzi, B.; Al-Rfou, R.; and Skiena, S. 2014.
\newblock Deepwalk: Online learning of social representations.
\newblock In \emph{Proceedings of the 20th ACM SIGKDD international conference
  on Knowledge discovery and data mining (KDD)}, 701--710.

\bibitem[{Poole et~al.(2019)Poole, Ozair, Van Den~Oord, Alemi, and
  Tucker}]{poole2019variational}
Poole, B.; Ozair, S.; Van Den~Oord, A.; Alemi, A.; and Tucker, G. 2019.
\newblock On variational bounds of mutual information.
\newblock In \emph{International Conference on Machine Learning (ICML)},
  5171--5180. PMLR.

\bibitem[{Qu et~al.(2021)Qu, Zhu, Zheng, Shi, and Yin}]{qu2021imgagn}
Qu, L.; Zhu, H.; Zheng, R.; Shi, Y.; and Yin, H. 2021.
\newblock ImGAGN: Imbalanced Network Embedding via Generative Adversarial Graph
  Networks.
\newblock In \emph{Proceedings of the 27th ACM SIGKDD Conference on Knowledge
  Discovery and Data Mining (KDD)}.

\bibitem[{Shchur et~al.(2018)Shchur, Mumme, Bojchevski, and
  G{\"u}nnemann}]{shchur2018pitfalls}
Shchur, O.; Mumme, M.; Bojchevski, A.; and G{\"u}nnemann, S. 2018.
\newblock Pitfalls of graph neural network evaluation.
\newblock \emph{arXiv preprint arXiv:1811.05868}.

\bibitem[{Shi et~al.(2020)Shi, Tang, Zhu, Wilson, and Liu}]{shi2020multi}
Shi, M.; Tang, Y.; Zhu, X.; Wilson, D.; and Liu, J. 2020.
\newblock Multi-class imbalanced graph convolutional network learning.
\newblock In \emph{Proceedings of the Twenty-Ninth International Joint
  Conference on Artificial Intelligence (IJCAI-20)}.

\bibitem[{Sun et~al.(2020)Sun, Hoffmann, Verma, and Tang}]{sun2019infograph}
Sun, F.-Y.; Hoffmann, J.; Verma, V.; and Tang, J. 2020.
\newblock Infograph: Unsupervised and semi-supervised graph-level
  representation learning via mutual information maximization.
\newblock In \emph{Proceedings of the International Conference on Learning
  Representations (ICLR)}.

\bibitem[{Tao et~al.(2021)Tao, Wang, Zhu, Dong, Song, Huang, and
  Dai}]{tao2021exploring}
Tao, C.; Wang, H.; Zhu, X.; Dong, J.; Song, S.; Huang, G.; and Dai, J. 2021.
\newblock Exploring the Equivalence of Siamese Self-Supervised Learning via A
  Unified Gradient Framework.
\newblock In \emph{IEEE Conference on Computer Vision and Pattern Recognition
  (CVPR)}.

\bibitem[{Thakoor et~al.(2022)Thakoor, Tallec, Azar, Munos,
  Veli{\v{c}}kovi{\'c}, and Valko}]{thakoor2021bootstrapped}
Thakoor, S.; Tallec, C.; Azar, M.~G.; Munos, R.; Veli{\v{c}}kovi{\'c}, P.; and
  Valko, M. 2022.
\newblock Large-Scale Representation Learning on Graphs via Bootstrapping.
\newblock In \emph{International Conference on Learning Representations
  (ICLR)}.

\bibitem[{Van~der Maaten and Hinton(2008)}]{van2008visualizing}
Van~der Maaten, L.; and Hinton, G. 2008.
\newblock Visualizing data using t-SNE.
\newblock \emph{Journal of machine learning research}, 9(11).

\bibitem[{Veli{\v{c}}kovi{\'c} et~al.(2018)Veli{\v{c}}kovi{\'c}, Fedus,
  Hamilton, Li{\`o}, Bengio, and Hjelm}]{velivckovic2018deep}
Veli{\v{c}}kovi{\'c}, P.; Fedus, W.; Hamilton, W.~L.; Li{\`o}, P.; Bengio, Y.;
  and Hjelm, R.~D. 2018.
\newblock Deep graph infomax.
\newblock In \emph{Proceedings of the International Conference on Learning
  Representations (ICLR)}.

\bibitem[{Wang, Aggarwal, and Derr(2021)}]{wang2021distance}
Wang, Y.; Aggarwal, C.; and Derr, T. 2021.
\newblock Distance-wise Prototypical Graph Neural Network in Node Imbalance
  Classification.
\newblock \emph{arXiv preprint arXiv:2110.12035}.

\bibitem[{Wei et~al.(2021)Wei, Sohn, Mellina, Yuille, and Yang}]{wei2021crest}
Wei, C.; Sohn, K.; Mellina, C.; Yuille, A.; and Yang, F. 2021.
\newblock Crest: A class-rebalancing self-training framework for imbalanced
  semi-supervised learning.
\newblock In \emph{Proceedings of the IEEE/CVF Conference on Computer Vision
  and Pattern Recognition (CVPR)}, 10857--10866.

\bibitem[{Xu et~al.(2021)Xu, Cheng, Luo, Chen, and Zhang}]{xu2021infogcl}
Xu, D.; Cheng, W.; Luo, D.; Chen, H.; and Zhang, X. 2021.
\newblock InfoGCL: Information-Aware Graph Contrastive Learning.
\newblock \emph{Advances in Neural Information Processing Systems (NeurIPS)},
  34.

\bibitem[{Yang and Xu(2020)}]{yang2020rethinking}
Yang, Y.; and Xu, Z. 2020.
\newblock Rethinking the value of labels for improving class-imbalanced
  learning.
\newblock In \emph{Thirty-Fourth Advances in Neural Information Processing
  Systems (NeurIPS)}.

\bibitem[{You et~al.(2020)You, Chen, Sui, Chen, Wang, and Shen}]{you2020graph}
You, Y.; Chen, T.; Sui, Y.; Chen, T.; Wang, Z.; and Shen, Y. 2020.
\newblock Graph contrastive learning with augmentations.
\newblock In \emph{Advances in neural information processing systems
  (NeurIPS)}.

\bibitem[{Zhang et~al.(2021{\natexlab{a}})Zhang, Wu, Yan, Wipf, and
  Yu}]{zhang2021canonical}
Zhang, H.; Wu, Q.; Yan, J.; Wipf, D.; and Yu, P.~S. 2021{\natexlab{a}}.
\newblock From canonical correlation analysis to self-supervised graph neural
  networks.
\newblock In \emph{Advances in Neural Information Processing Systems
  (NeurIPS)}.

\bibitem[{Zhang et~al.(2021{\natexlab{b}})Zhang, Kang, Hooi, Yan, and
  Feng}]{zhang2021deep}
Zhang, Y.; Kang, B.; Hooi, B.; Yan, S.; and Feng, J. 2021{\natexlab{b}}.
\newblock Deep long-tailed learning: A survey.
\newblock \emph{arXiv preprint arXiv:2110.04596}.

\bibitem[{Zhang et~al.(2021{\natexlab{c}})Zhang, Wei, Zhou, and
  Wu}]{zhang2021bag}
Zhang, Y.; Wei, X.-S.; Zhou, B.; and Wu, J. 2021{\natexlab{c}}.
\newblock Bag of Tricks for Long-Tailed Visual Recognition with Deep
  Convolutional Neural Networks.
\newblock In \emph{Proceedings of the AAAI Conference on Artificial
  Intelligence}, volume~35, 3447--3455.

\bibitem[{Zhao, Zhang, and Wang(2021)}]{zhao2021graphsmote}
Zhao, T.; Zhang, X.; and Wang, S. 2021.
\newblock GraphSMOTE: Imbalanced Node Classification on Graphs with Graph
  Neural Networks.
\newblock In \emph{Proceedings of the 14th ACM International Conference on Web
  Search and Data Mining (WSDM)}, 833--841.

\bibitem[{Zhu et~al.(2021{\natexlab{a}})Zhu, Xu, Liu, and
  Wu}]{zhu2021empirical}
Zhu, Y.; Xu, Y.; Liu, Q.; and Wu, S. 2021{\natexlab{a}}.
\newblock An Empirical Study of Graph Contrastive Learning.
\newblock In \emph{Thirty-fifth Conference on Neural Information Processing
  Systems Datasets and Benchmarks Track}.

\bibitem[{Zhu et~al.(2021{\natexlab{b}})Zhu, Xu, Yu, Liu, Wu, and
  Wang}]{zhu2020graph}
Zhu, Y.; Xu, Y.; Yu, F.; Liu, Q.; Wu, S.; and Wang, L. 2021{\natexlab{b}}.
\newblock Graph Contrastive Learning with Adaptive Augmentation.
\newblock In \emph{Proceedings of the Web Conference (WWW)}, 2069--2080.

\end{thebibliography}

\clearpage
\appendix
\section{Theoretical Analysis}
In order to justify our proposed method, we theoretically prove that, on the imbalanced unlabeled dataset, a classifier learned iteratively by balanced sampling converges to the optimal balanced classifier with a linear rate.

Consider a binary classification problem with two Gaussian distributions with different means and the equal variance. Suppose the data generating distribution is $P_{XY}$ and the probability of positive labels $(+1)$ and negative labels $(-1)$ are $P_Y(1)$ and $P_Y(-1)$, respectively. We have $X|Y=+1 \sim \mathcal{N}(\mu_1, \sigma^2)$ conditioned on $Y=+1$ and similarly, $X|Y=-1 \sim \mathcal{N}(\mu_2, \sigma^2)$ conditioned on $Y=-1$. In addition, suppose $\mu_1 < \mu_2$ without loss of generality. We consider the imbalanced setting and it is straightforward to have~\cite{bishop2006pattern}:
\begin{fact}
\label{coro1}
Consider the above setup, the optimal Bayes classifier is
% \begin{align}
%     f(x)
%     % &=\text{sign} \left(x - \frac{\mu_1+\mu_2}{2} - \frac{\sigma^2}{\mu_2-\mu_1}\log \frac{P_Y(1)}{P_Y(-1)}\right),
%     % \text{sign}\left(\left(\vmu_2^T-\vmu_1^T\right)\mSigma^{-1}\left(\vx - \frac{\vmu_1+\vmu_2}{2}\right) - \log \frac{P_Y(1)}{P_Y(-1)}\right)\\
%     % \text{sign} \left(\left(\bm{x} - \frac{\bm{\mu_1}+\bm{\mu_2}}{2}\right)\cdot(\bm{\mu_2}-\bm{\mu_1}) - \frac{1}{2}\log \frac{P_Y(1)}{P_Y(-1)}\right)\\
%     &= \text{sign} \left(x - \theta^* -  \frac{\sigma^2}{\mu_2-\mu_1}\log \frac{P_Y(1)}{P_Y(-1)}\right)\label{eq:optim_bayes_classifier}
%     % (\textup{1d case}),
% \end{align}
$f(x) = \text{sign} \left(x - \theta^* -  \frac{\sigma^2}{\mu_2-\mu_1}\log \frac{P_Y(1)}{P_Y(-1)}\right)\label{eq:optim_bayes_classifier}$ with the optimal balanced decision boundary $\theta^*$ being
$\theta^* \equiv \frac{\mu_1+\mu_2}{2}$, where $\text{sign}(x)$ is the indicator function with $\text{sign}(x) = 1$ if $x>0$ and $-1$ otherwise.
% $\text{sign}(x) = -1$ otherwise.
\end{fact}

\paragraph{Interpretation.} Intuitively, if the label $Y$ is either positive $(+1)$ or negative $(-1)$ with equal probability, the optimal Bayes's classifier boundary is the perpendicular hyperplane through the midpoint of two Gaussian centers. However, if the number of positive labels is larger than that of negative labels, the decision boundary will be dominated by the positive~(head) class and shift towards the negative~(tail) class.

In the context of imbalanced learning, the training dataset is imbalanced but the testing dataset is balanced. However, in self-supervised GCL, label information is absent and we instead propose to create pseudo-labels iteratively via representations learned from GCL to enable the PBS method. 
% At the first iteration $t=1$, one starts from the imbalanced unlabeled dataset and generates pseudo-labels $\hat{Y}_1$ by the learned decision boundary estimator $\hat{\theta}_1$. 
At the first iteration $t=1$, one starts from the imbalanced unlabeled dataset and generates pseudo-labels $\hat{Y}_0$ by the PBS method. Then, we obtain the learned decision boundary estimator $\hat{\theta}_1$ and produce pseudo-labels $\hat{Y}_1$. 
The fact that the initial dataset being imbalanced (\ie, $P_Y(1)\neq P_Y(-1)$) leads to biased $\hat{\theta}_1$ and $\hat{f}_1$ by Eq.~\ref{eq:optim_bayes_classifier}, which has three consequences: (1) The learned classifier $\hat{f}_1$ exhibits higher recall/lower precision for the head classes and the opposite for the tail classes~\cite{yang2020rethinking}, aligned with our empirical observations in Fig.~\ref{fig:per-class}. (2) For any iteration $t$, the biased classifier $\hat{f}_t$ and pseudo-labels $\hat{Y}_t$ induce distribution shift between the pseudo-classes $\hat{X}$ and ground truth classes $X$. (3) For any following iteration $t$, the dataset re-balanced based on $\hat{Y}_t$ is still imbalanced with respect to the ground truth labels $Y$, which by Eq.~\ref{eq:optim_bayes_classifier} means $|\hat{\theta}_{t+1} - \theta^*| > 0, \forall t \in\mathbb{N}_{+}$. For which we prove the following proposition:

\begin{prop}
\label{prop1}
Consider the above setup. Suppose there is a (sufficiently large) integer 
$T$ such that $|\hat{\theta}_T-\theta^*|\ll |\mu_2-\mu_1|$, $|(\hat{\theta}_T-\theta^*)(\mu_2-\mu_1)|\ll\sigma^2$.
% for all $t>T$, we have $\theta_t-\theta^*=\delta_t$ $(\delta_t \rightarrow 0^+)$. 
Our estimator converges to the optimal balanced decision boundary $\theta^*$, i.e. $\left|\hat{\theta}_{t+1} - \theta^* \right| \leq C \cdot \left|\hat{\theta}_{t} - \theta^* \right|, \forall t\ge T$ with a linear convergence rate $C = \frac{2}{\pi} < 1$.
% then satisfies $\left|\hat{\theta}_{t+1} - \theta^* \right| \leq C \left|\hat{\theta}_{t} - \theta^* \right|$ with linear convergence rate $C < 1$. 
% Proof. \textup{ See Supplement~\ref{sec:appendix-A}.}
\end{prop}

\paragraph{Interpretation.} 
In the context of PBS, the classifier is trained starting from the imbalanced data distribution. Intuitively, by iteratively down-sampling head classes, the data distribution gradually becomes balanced, on which the trained classifier also converges to the balanced optimum.
% imbalanced learning, we have the imbalanced data distribution for the training dataset but balanced data distribution for the testing dataset, \textit{i.e.}, $P_Y(1)/P_Y(-1)$ is substantially larger~(or less) than $1$ in the training dataset, however $P_Y(1)/P_Y(-1)$ is equal to $1$ in the testing dataset.
As stated in proposition~\ref{prop1}, the classifier learned by balanced sampling with pseudo-label on the imbalanced dataset can converge to the optimal balanced classifier with a linear rate, which justifies the rationale of the PBS method.
% \section*{supplemental material}
% \setcounter{page}{1}
\section{Mathematics Proof}
\label{sec:appendix-A}
\subsection{Proof of Fact 1}
The optimal Bayes's classifier is the probabilistic model that makes the most probable prediction for new examples. Thus, we have:
\begin{equation}
\begin{aligned}
    f^*(x) &= \arg\max_{y} P_{Y|X}(y|x) \\
    &= \arg\max_{y} P_{X|Y}(x|y) P_Y(y) \;,
\end{aligned}
\end{equation}
where we omit $P_X(x)$ in the denominator, which is unrelated to the objective function. Due to the monotonicity property of the \emph{log} function, we have:
\begin{equation}
\begin{aligned}
    f^*(x) &= \arg\max_{y} \{\log P_{X|Y}(x|y) + \log P_Y(y)\} \\
    &= \arg\max_{y} \{ \log[ \frac{1}{\sqrt{2\pi\sigma^2}} \exp{\{-\frac{(x-\mu_y)^2}{2\sigma^2}\}} ] +\\
    &\qquad\qquad\quad \log P_Y(y) \} \\
    % &= \arg\max_{y} \left\{ -\frac{1}{2}\log(2\pi\sigma^2) - \frac{(x-\mu_y)^2}{2\sigma^2} + \log P_Y(y) \right\} \\
    &= \arg\min_{y} \left\{ \frac{(x-\mu_y)^2}{2\sigma^2} - \log P_Y(y) \right\} \\
    % &= \arg\min_{y} \left\{ x^2 - 2x\mu_y + \mu_y^2 - 2\sigma^2\log P_Y(y) \right\} \\
    &= \arg\min_{y} \left\{ - 2x\mu_y + \mu_y^2 - 2\sigma^2\log P_Y(y) \right\} .
\end{aligned}
\end{equation}
So, we will pick the positive sample $+1$ if
\begin{equation}
\begin{aligned}
    &-2x\mu_1 + \mu_1^2 - 2\sigma^2\log P_Y(1) < -2x\mu_2 + \mu_2^2 \\
    &\qquad\qquad\qquad\qquad\qquad\qquad\qquad\quad - 2\sigma^2\log P_Y(-1) \\
    &\Rightarrow 2x(\mu_2-\mu_1) < \mu_2^2 - \mu_1^2 + 2\sigma^2\log\frac{P_Y(1)}{P_Y(-1)} \\
    &\qquad\qquad\qquad\qquad\qquad\qquad\qquad (\text{because of} \; \mu_1 < \mu_2)\\
    &\Rightarrow x < \frac{\mu_1+\mu_2}{2} + \frac{\sigma^2}{\mu_2-\mu_1} \log\frac{P_Y(1)}{P_Y(-1)} .
\end{aligned}
\end{equation}
Therefore, the optimal Bayes's classifier is:
\begin{equation}
    f^*(x) = \text{sign}\left(x - \frac{\mu_1+\mu_2}{2} - \frac{\sigma^2}{\mu_2-\mu_1} \log\frac{P_Y(1)}{P_Y(-1)}\right) \;,
\end{equation}
where $\text{sign}$ is the indicator function with $\text{sign}(x) = 1$ if $x>0$ and $\text{sign}(x) = -1$ otherwise.

\subsection{Proof of Proposition 1}
At training step $t$, we obtain the biased decision boundary $\hat{\theta}_t$ and generate corresponding pseudo-labels $\hat{Y}_t$. We perform the balanced sampling method on $\hat{Y}_t$ to get balanced positive and negative samples and then take them as input for the next training step to get $\hat{\theta}_{t+1}$. Suppose the probability of true labels on selected positive samples~$(+1)$ and negative samples~$(-1)$ are $P_{Y,t}(1)$ and $P_{Y,t}(-1)$, respectively. According to Fact~\ref{coro1}, the decision boundary $\hat{\theta}_t$ produces the bias term $\Delta = \frac{\sigma^2}{\mu_2-\mu_1} \log\frac{P_{Y,t}(1)}{P_{Y,t}(-1)}$. We intend to prove that, at training step $t+1$, we get a smaller bias term $C\Delta$, where $C<1$. Without loss of generality, we set $\sigma=1$ 
% without loss of generality 
in what follows, which can be recovered by dimensional analysis. 
The optimal balanced decision boundary is $\theta^* = \frac{\mu_1+\mu_2}{2}$, and we construct the estimator $\hat{\theta}_{t+1}=\frac{\hat{\mu}_{t,1}+\hat{\mu}_{t,2}}{2}$ by the balanced sampling method. Therefore, we need to estimate $\hat{\mu}_{t,1}$ and $\hat{\mu}_{t,2}$ conditional on the decision boundary $\hat{\theta_t}$. 
Suppose the cumulative distribution function~(CDF) of the standard normal distribution with a mean of $0$ and standard deviation of $1$ is $\Phi(\cdot)$.
For clarity, we denote $x=\frac{\mu_2-\mu_1}{2} > 0$ assuming $\mu_1 < \mu_2$ and $A=\frac{1}{\sqrt{2\pi}}e^{-\frac{x^2}{2}}$. The imbalanced ratio of positive and negative samples is defined as:
\begin{equation}
    \frac{p_{Y,t}(1)}{p_{Y,t}(-1)} = e^{\Delta(\mu_2-\mu_1)} .
    \label{eq:imbalance_ratio}
\end{equation}
Since $\Delta = |\theta_t-\theta^*|$ satisfies $\Delta\ll \left|\mu_2-\mu_1\right|$ and $|\Delta(\mu_2-\mu_1)|\ll\sigma^2=1$, we have the following equation by first-order approximation:
\begin{equation}
\begin{aligned}
    e^{-(x+\Delta)^2} &\approx e^{-x^2} - 2x e^{-x^2}*\Delta \\
    \Phi(x+\Delta) &\approx \Phi(x) + \Phi'(x)*\Delta \\
    &= \Phi(x) + \frac{1}{\sqrt{2\pi}}e^{-\frac{x^2}{2}}*\Delta .
\end{aligned}
\end{equation}
The probability of $Y=1$ on the left of the decision boundary $\hat{\theta}_t$ is:
\begin{equation}
\begin{aligned}
    \Phi\left(\frac{\mu_1+\mu_2}{2}-\mu_1+\Delta\right)P_{Y,t}(1) &= \Phi\left(\frac{\mu_2-\mu_1}{2}+\Delta\right)P_{Y,t}(1) \\
    &= \Phi(x+\Delta)P_{Y,t}(1).
\end{aligned}
\end{equation}
The probability of $Y=-1$ on the left of the decision boundary $\hat{\theta}_t$ is:
\begin{equation}
    \Phi\left(\frac{\mu_1+\mu_2}{2}-\mu_2+\Delta\right)P_{Y,t}(-1) = \Phi(-x+\Delta)P_{Y,t}(-1) .
\end{equation}
The total probability mass on the left of the decision boundary $\hat{\theta}_t$ is:
\begin{align}
    D_{L} &= \Phi(x+\Delta)P_{Y,t}(1) + \Phi(-x+\Delta)P_{Y,t}(-1)\\
    &= [\Phi(x)+A\Delta]P_{Y,t}(1) + [\Phi(-x)+A\Delta]P_{Y,t}(-1).
\end{align}
By translating the Gaussian to the zero mean, the unnormalized mean of $Y=1$ on the left of the decision boundary $\hat{\theta}_t$ is:
\begin{equation}
\begin{aligned}
    &\left[\mu_1\Phi(x+\Delta) + \frac{1}{\sqrt{2\pi}}\int_{-\infty}^{x+\Delta}x e^{-\frac{x^2}{2}}dx\right]P_{Y,t}(1) \\
    &=\left[\mu_1\Phi(x+\Delta) - \frac{1}{\sqrt{2\pi}}e^{-\frac{(x+\Delta)^2}{2}}\right]P_{Y,t}(1) \\
    &= \left[\mu_1\left(\Phi(x)+A\Delta\right) - A + Ax\Delta \right]P_{Y,t}(1) .
\end{aligned}
\end{equation}
Similarly, the unnormalized mean of $Y=-1$ on the left of the decision boundary $\hat{\theta}_t$ is:
\begin{equation}
\begin{aligned}
    &\left[\mu_2\Phi(-x+\Delta) + \frac{1}{\sqrt{2\pi}}\int_{-\infty}^{-x+\Delta}x e^{-\frac{x^2}{2}}dx\right]P_{Y,t}(-1) \\
    &=\left[\mu_2\Phi(-x+\Delta) - \frac{1}{\sqrt{2\pi}}e^{-\frac{(-x+\Delta)^2}{2}}\right]P_{Y,t}(-1) \\
    &= \left[\mu_2\left(\Phi(-x)+A\Delta)\right) - A - Ax\Delta \right]P_{Y,t}(-1) .
\end{aligned}
\end{equation}
The estimated mean on the left of the decision boundary is given by $N_L/D_L$ where:
\begin{equation}
\begin{aligned}
    N_L &= \left[\mu_1\left(\Phi(x)+A\Delta)\right) - A + Ax\Delta \right]P_{Y,t}(1) \\
    &+ \left[\mu_2\left(\Phi(-x)+A\Delta)\right) - A - Ax\Delta \right]P_{Y,t}(-1) .\\
\end{aligned}
\end{equation}
According to Eq.~\ref{eq:imbalance_ratio}, we have $\frac{p_{Y,t}(1)}{p_{Y,t}(-1)} = e^{\Delta(\mu_2-\mu_1)} \approx 1+\Delta(\mu_2-\mu_1)$. Dividing both $N_L$ and $D_L$ by $P_{Y,t}(-1)$, we have:
\begin{equation}
\begin{aligned}
    D_L &= \left[\Phi(x)+A\Delta\right]\left[1+\Delta(\mu_2-\mu_1)\right] + \left[\Phi(-x) + A\Delta\right] \\
    N_L &= \left[\mu_1\left(\Phi\left(x\right)+A\Delta\right)-A+Ax\Delta\right] \left[1+\Delta\left(\mu_2-\mu_1\right)\right] \\
    &+ \left[\mu_2\left(\Phi\left(-x\right)+A\Delta\right)-A-Ax\Delta\right] .
    \label{eq:DLNL}
\end{aligned}
\end{equation}
Similarly, on the right of the decision boundary, we get:
\begin{equation}
\begin{aligned}
    D_R &= \left[\Phi(-x)-A\Delta\right]\left[1+\Delta(\mu_2-\mu_1)\right] + \left[\Phi(x) - A\Delta\right] \\
    N_R &= \left[\mu_1\left(\Phi\left(-x\right)-A\Delta\right)+A-Ax\Delta\right]\left[1+\Delta\left(\mu_2-\mu_1\right)\right] \\
    &+ \left[\mu_2\left(\Phi\left(-x\right)-A\Delta\right)+A+Ax\Delta\right] .
    \label{eq:DRNR}
\end{aligned}
\end{equation}
Next, we can estimate $\hat{\mu}_{t,1}$ and $\hat{\mu}_{t,2}$ according to Eq.~\ref{eq:DLNL} and~\ref{eq:DRNR}.
\begin{equation}
\scriptsize
\begin{aligned}
    \hat{\mu}_{t,1} = \frac{N_L}{D_L} & = \frac{[\mu_1\Phi(x)-2A+\mu_2\Phi(-x)]+\Delta\mu_1\left[2A+\left(\mu_2-\mu_1\right)\Phi(x)\right]}{1+\Delta\left[2A+\left(\mu_2-\mu_1\right)\Phi(x)\right]} \\
    &= \left\{[\mu_1\Phi(x)-2A+\mu_2\Phi(-x)]+\Delta\mu_1\left[2A+\left(\mu_2-\mu_1\right)\Phi(x)\right]\right\}\\ &*\left\{1-\Delta\left[2A+\left(\mu_2-\mu_1\right)\Phi(x)\right]\right\} \\
    &= [\mu_1\Phi(x)-2A+\mu_2\Phi(-x)]\\
    &+ \Delta \left[2A+\left(\mu_2-\mu_1\right)\Phi(x)\right] \left[2A-\left(\mu_2-\mu_1\right)\Phi(-x)\right] ,
\end{aligned}
\end{equation}
\begin{equation}
\scriptsize
\begin{aligned}
    \hat{\mu}_{t,2} = \frac{N_R}{D_R} & = \frac{[\mu_1\Phi(-x)+2A+\mu_2\Phi(x)]+\Delta\mu_1\left[-2A+\left(\mu_2-\mu_1\right)\Phi(-x)\right]}{1+\Delta\left[-2A+\left(\mu_2-\mu_1\right)\Phi(-x)\right]} \\
    &= \left\{[\mu_1\Phi(-x)+2A+\mu_2\Phi(x)]+\Delta\mu_1\left[-2A+\left(\mu_2-\mu_1\right)\Phi(-x)\right]\right\} \\
    &*\left\{1-\Delta\left[-2A+\left(\mu_2-\mu_1\right)\Phi(-x)\right]\right\} \\
    &= [\mu_1\Phi(-x)+2A+\mu_2\Phi(x)] \\
    &+ \Delta \left[2A-\left(\mu_2-\mu_1\right)\Phi(-x)\right] \left[2A+\left(\mu_2-\mu_1\right)\Phi(x)\right] ,
\end{aligned}
\end{equation}
where we use the property of the $\Phi$ function, \ie, $\Phi(x)=1-\Phi(-x)$. Then, we have:
\begin{equation}
\begin{aligned}
    \hat{\theta}_{t+1}&=\frac{\hat{\mu}_{t,1}+\hat{\mu}_{t,2}}{2}=\frac{1}{2}(\frac{N_L}{D_L} + \frac{N_R}{D_R}) \\
    &= \frac{\mu_1+\mu_2}{2} + \\
    &\Delta \left[2A+\left(\mu_2-\mu_1\right)\Phi(x)\right] \left[2A-\left(\mu_2-\mu_1\right)\Phi(-x)\right] .
\end{aligned}
\end{equation}
Finally, we can get:
\begin{equation}
\begin{aligned}
    C &= \left[2A+\left(\mu_2-\mu_1\right)\Phi(x)\right] \left[2A-\left(\mu_2-\mu_1\right)\Phi(-x)\right] \\
    &= (2A+2x\Phi(x))(2A-2x\Phi(-x)) \\
    &= 4(A+x\Phi(x))(A-x(1-\Phi(x))) \\
    &= 4\left(A^2+Ax(2\Phi(x)-1)-x^2\Phi(x)(1-\Phi(x))\right) ,x\geq 0.
\end{aligned}
\end{equation}
We can verify that 
% the derivative of the RHS is 
% \begin{equation}
%     C' = 4\left[A(2\Phi(x)-1)-2x\Phi(x)(1-\Phi(x))\right] < 0, \forall x\ge 0
% \end{equation}
when $x=0$, one obtains the maximal value $C = 4A^2=\frac{2}{\pi} \approx 0.637 < 1$. 
This completes the proof.% \rightline{$\qed$}

\section{Ablation Study on Amazon-Computers}
We conduct ablations using the GBT baseline on Amazon-computers due to its largest imbalanced factor among four datasets, as shown in Table~\ref{table:ablation}. In the imbalanced node classification setting, the performance for the head, middle, and tail classes are usually reported~\cite{zhang2021deep}.
We first divide the Amazon-computers dataset into three disjoint groups in terms of class size: \{\emph{Head}, \emph{Middle}, \emph{Tail}\}. \emph{Head} and \emph{Tail} each include the top and bottom $1/3$ classes, respectively. Because there are 10 classes of Amazon-computers in total, the classes with sorted indices in decreasing order [1-3, 4-7, 8-10] belong to [\emph{Head}~(3 classes), \emph{Middle}~(4 classes), \emph{Tail}~(3 classes)] categories, respectively.
\begin{table*}
\small
\begin{tabular}{lcccc}
\toprule
                                   & \textit{Many}        & \textit{Medium}       & \textit{Few}          & All         \\ \midrule
GBT                                & $70.38_{\pm5.96}$ & $41.92_{\pm3.28}$  & $14.29_{\pm3.53}$    & $42.17_{\pm3.74}$ \\
\quad+\footnotesize{K-means}                               & $52.09_{\pm9.44}$ & $84.16_{\pm8.82}$  & $89.00_{\pm10.20}$ & $75.99_{\pm6.75}$ \\
\quad+\footnotesize{Community detection}                         & $51.22_{\pm8.30}$ & $84.61_{\pm8.64}$   & $91.20_{\pm9.22}$  & $76.57_{\pm3.87}$ \\
\quad+\footnotesize{ImGCL w/o node centrality}    & $56.45_{\pm9.13}$ & $\bm{86.08_{\pm9.30}}$  & $84.44_{\pm6.49}$  & $76.70_{\pm6.94}$ \\
\rowcolor{black!10}\quad+\footnotesize{ImGCL (ours)} & $57.32_{\pm7.29}$ & $82.72_{\pm8.46}$  & $\bm{94.45_{\pm6.07}}$  & $78.62_{\pm1.73}$ \\
\quad+\footnotesize{True labels}                       & $\bm{58.43_{\pm8.37}}$ & $84.40_{\pm10.55}$ & $92.24_{\pm10.60}$ & $\bm{78.96_{\pm2.92}}$ \\ \midrule
$\textbf{GBT over GBT+ImGCL}$                              & \color{blue}\textbf{13.06  $\downarrow$}       & \color{red}\textbf{40.80  $\uparrow$}        & \color{red}\textbf{80.16  $\uparrow$}        & \color{red}\textbf{36.45  $\uparrow$}       \\ \bottomrule
\end{tabular}
\centering\caption{Ablation studies on the proposed components of ImGCL. We conduct experiments on Amazon-Computers based on the GBT model. \emph{Many}, \emph{Medium}, and \emph{Few} are split according to the number of classes in decreasing order.}
% Details can be found in Sec.~\ref{sec:exp_ablation}.
\label{table:ablation}
\end{table*}
We consider four variants of GBT+ImGCL in three settings: GBT with one-stage pseudo-labels (k-means and community detection), multi-stage pseudo-labels (PBS without node centrality), and ground truth labels (true labels).
The ``k-means'' and ``community detection'' variants only consider node features and graph topology respectively.
Compared with these two variants, we conclude that graph topology is more important for PBS of GCL. 
This aligns with the observation that PBS using node centrality information (GBT+ImGCL) significantly outperforms ordinary PBS (GBT+PBS), which also shows that the proposed node centrality~(PageRank centrality) is an effective component in PBS. 
Moreover, ImGCL achieves better or comparable results compared to the variant with true labels, which means that ImGCL indeed generates faithful pseudo-labels from the learned representations in an iterative manner. Overall, ImGCL substantially outperforms the baseline in terms of overall accuracy and improves the performance on tail and middle classes by a large margin at cost of slightly sacrificing the accuracy on head classes.
\section{Hyperparameter Study}
\label{sec:exp_hyperparameter}
We study the hyperparameter sensitivity of our proposed framework ImGCL, and conduct experiments on Amazon-Computers based on the GBT model. We have two main hyperparameters in ImGCL: the number of clusters and the re-balanced labeling frequency $B$. 
The \#Clusters denotes the number of predefined clusters which are equivalent to the classes of pseudo-labels.
The \#Labeling $=T/B$ denotes the number of stages in which class distribution is re-balanced based on newly-generated pseudo-labels. 
We can observe that, when \#Clusters equals the classes of downstream task~(10 in Amazon-Computers), ImGCL achieves the highest accuracy~($78.62_{\pm1.73}$). These observations are in line with the fact that when the semantic classes of learned representations in ImGCL reconcile with the actual classes in the downstream task, ImGCL obtains the highest performance gain from the self-supervised GCL methods.
Moreover, when \#Clusters is less than 8, the performance fluctuates violently. We think the reason is that the learned representations in ImGCL could degenerate into few centroids when the number of predefined clusters is relatively small.
As the \#Labeling varies from $6$ to $12$, our proposed ImGCL achieves relatively stable performance. We can also find that when \#Labeling is larger than 10, the accuracy slightly decreases but the variance largely increases. We conjecture that when we produce pseudo-labels frequently, the GCL model cannot quickly adapt to the new sample distribution, which leads to large fluctuations in performance.
In summary, we find the performance of our framework is relatively stable across different values of \#Clusters and \#Labeling hyperparameters. Thus, ImGCL does not rely on heavy and case-by-case hyperparameter tuning to achieve satisfactory results.
\begin{figure}[t]
    \centering
    \includegraphics[width=1.\linewidth]{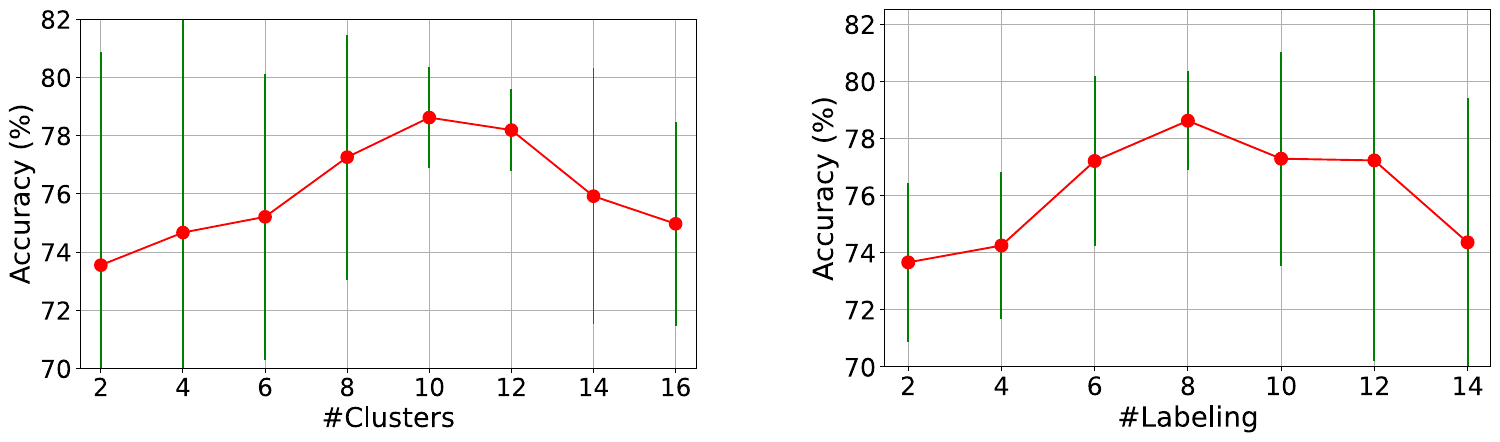}
    % \vspace{-1em}
    \caption{Hyperparameter study on Amazon-Computers.}
    % \vspace{-1em}
    \label{fig:hyper-parameter}
\end{figure}
\section{Computational Complexity}

ImGCL is a general GCL framework. Suppose we use the traditional GCN~\cite{kipf2016semi} model in GCL, the training time complexity of a GCN with $L$ layers is $\mathcal{O}\left(Lmd+Lnd^2\right)$, where $d$ is the dimension of the learned representations, and $n=|\gV|$ and $m=|\gE|$ denote the number of vertices and edges in the given graph, respectively. The time complexity of the clustering algorithm is $\mathcal{O}\left(nd\right)$ in ImGCL. We need $k=T/B$ times to adjust the class distribution using the clustering algorithm to produce pseudo-labels, thus we have an extra $\mathcal{O}\left(knd\right)$ time complexity. 
Note $k$ is always less than 10 in our experiments, so the time of the clustering algorithm in ImGCL is negligible compared with the training of GNN encoders. Moreover, the time complexity of calculating node centrality is $\mathcal{O}\left(t*n\right)$, where $t$ is the number of iterations in PageRank. This step can be seen as a pre-computing step.
\section{More Discussions on Graph Contrastive Learning and Imbalanced Learning}
\label{sec:related_work}
\noindent \textbf{Graph Contrastive Learning.} 
Some recent remarkable progress has been made to adapt contrastive learning methods to the graph domain. Typically, the main idea behind graph contrastive learning~(GCL) is to learn effective representations in which similar sample pairs stay close to each other while dissimilar ones are far apart~\cite{zhu2021empirical}. According to~\cite{tao2021exploring}, we can divide GCL methods into three categories: (1) \emph{contrastive learning} methods~\cite{velivckovic2018deep, sun2019infograph, hassani2020contrastive, you2020graph, zhu2020graph} aim to reduce the distance between two augmented views of the same nodes/graphs~(positive samples), and push apart views of different nodes/graphs~(negative samples). These methods mainly differ in several design choices including graph data augmentations, contrasting modes, and contrastive objectives. DGI~\cite{velivckovic2018deep} adopts the InfoNCE principle~\cite{poole2019variational} to the node classification task by contrasting the node- and graph-level representations of the given graph. InfoGraph~\cite{sun2019infograph} mainly focus on graph classification task by contrasting graph representation of different granularity via mutual information maximization. MVGRL~\cite{hassani2020contrastive} utilizes the graph diffusion matrix to construct the augmentation views of the original graph and then contrast node- and graph-level representation by scoring the agreement between representations. GraphCL~\cite{you2020graph} adopts the SimCLR~\cite{chen2020simpleframework} framework and propose four practical graph augmentation methods to obtain better graph representations. GRACE~\cite{zhu2020graph} capitalizes on the idea that data augmentations for GCL should preserve the property of graphs to highlight important node features.
(2) Negative samples play an important role in contrastive learning methods. However, \emph{asymmetric network} methods~\cite{thakoor2021bootstrapped} get rid of negative samples via the introduction of a predictor network and the stop-gradient operation to avoid mode collapse.
(3) \emph{Feature decorrelation} methods~\cite{zhang2021canonical, bielak2021graph} has recently been introduced as a new solution to GCL. They focus on reducing the redundancy among feature dimensions by introducing a loss function calculated by the cross-correlation matrix of two augmentation views.
Different from our work, these methods mainly assume that the unlabeled graph data distribution is always balanced for GCL. However, in practice, the (unlabeled) class distribution could be highly skewed and lead to sub-optimal generalization on the imbalanced graph dataset.

\noindent \textbf{Imbalanced Learning.} 
It is well known that the degree of many large graphs follows a scale-free power-law distribution, which is long-tailed and highly imbalanced~\cite{barabasi1999emergence}. 
In this work, we focus on the imbalanced class distribution, which is also ubiquitous for graph datasets. There exists some recent work to study the class imbalance of semi-supervised node classification problems. They mainly improve the imbalanced classification performance by re-balancing the sample ratio of head and tail classes~\cite{zhao2021graphsmote, qu2021imgagn, chawla2002smote} and transfer knowledge from head classes to enhance model performance on tail classes~\cite{liu2020towards, wang2021distance, shi2020multi}. GraphSMOTE~\cite{zhao2021graphsmote} utilizes the SMOTE synthetic over-sampling algorithm to up-sample the tail nodes and then improve their prediction performance. ImGAGN~\cite{qu2021imgagn} aims to improve the representation learning of tail nodes by generating adversarial tail nodes to balance the class distribution of the original graph. DPGNN~\cite{wang2021distance} borrows the idea of metric learning to transfer knowledge from head classes to tail classes to boost model performance. However, they mainly ignore that open-world unlabeled data usually follows the imbalanced distribution, which will lead to performance degradation under unsupervised or self-supervised learning. To our best knowledge, we are the first to explore contrastive learning on \emph{highly} imbalanced graph datasets.
\section{More Discussions on Data Sampling Strategies}

Different sampling strategies have different specific values of $q$ in Eq.~\ref{equ:sampling}, which have different practical meanings introduced below.

\noindent\textbf{Random Sampling.} The probability of a node from the class $k$ is defined as $p_{k}^{R} = \frac{N_k}{\sum_{i=1}^K N_i}$ by setting $q=1$ in Eq.~\ref{equ:sampling}, referred to as instance-balanced sampling in some literature~\cite{zhang2021deep}. A node from class $k$ is sampled proportionally according to the cardinality $N_k$ of each class. If we sample nodes from a given graph according to $p_{k}^{R}$, the highly imbalanced property still remains just like the training dataset in imbalanced learning. It reflects the highly imbalanced and skewed class distribution of real-world data collection.

\noindent\textbf{Mean Sampling.} The probability of a node from the class $k$ is given by $p_{k}^{M}=\frac{1}{K}$ by setting $q=0$ in Eq.~\ref{equ:sampling}, referred to as class-balanced sampling in some literature~\cite{zhang2021deep}. A node is sampled with an equal probability for each class. Sampling each class with an equal probability according to $p_{k}^{M}$ has been found to significantly improve the performance of the classifier in long-tailed learning, while being sub-optimal for representation learning~\cite{kang2019decoupling}.
\section{Experimental Setup}
\label{sec:appendix-B}
For reproducibility, we provide all websites of the baseline GNN models and datasets. Our codes and data are publicly available.
% ~\footnote{\url{https://tinyurl.com/ImGCL}}.

\subsection{Experimental Settings}
All experiments are conducted with the following settings:
\begin{itemize}[leftmargin=10pt]
    \item Operating system: Linux Red Hat 4.8.2-16
    \item CPU: Intel(R) Xeon(R) Platinum 8255C CPU @ 2.50GHz
    \item GPU: NVIDIA Tesla V100 SXM2 32GB
    \item Software versions: Python 3.8.10; Pytorch 1.9.0+cu102; Numpy 1.20.3; SciPy 1.7.1; Pandas 1.3.4; scikit-learn 1.0.1; PyTorch-geometric 2.0.2; DGL 0.7.2; Open Graph Benchmark 1.3.2
\end{itemize}

\subsection{Datasets}
The statistics of datasets we used in experiments are summarized in Table~\ref{table:stat}. We also provide a brief description of datasets in the experiment section as follows:
\begin{itemize}[leftmargin=10pt]
    \item \textbf{Wiki-CS}~\cite{zhu2021empirical} is a reference network constructed based on Wikipedia. The nodes represent articles about computer science and the edges represent the hyperlinks between two articles. Each node corresponds to a label representing a branch of the field. Node features are calculated by the average of word embeddings in each article.
    \item \textbf{Amazon-Computers} and \textbf{Amazon-Photo}~\cite{shchur2018pitfalls} are two networks of co-purchase relationships constructed from Amazon. The nodes represent goods and the edges indicate that two goods are frequently bought together. Node features are bag-of-words encoded product reviews, and class labels are given by the product category.
    \item \textbf{DBLP}~\cite{shchur2018pitfalls} is a citation network constructed from DBLP website. The nodes represent papers and the edges represent the citation relationship between two papers. For each paper, a sparse bag-of-words vector is utilized as the node feature vector. The node labels are defined according to their research topic.
\end{itemize}

\begin{table}[]
\small
\setlength{\tabcolsep}{5pt}
% \vspace{-1em}
\begin{tabular}{lcccccccc}
\toprule
                 & \makecell*[c]{Amazon-\\Computers} & \makecell*[c]{Amazon-\\Photo} & Wiki-CS & DBLP   \\ \midrule
\#Nodes          & 13752            & 7650         & 11701   & 17716  \\
\#Edges          & 245861           & 119081       & 216123  & 105734 \\
\#Features       & 767              & 745          & 300     & 1639   \\
\#Classes        & 10               & 8            & 10      & 4      \\
\#Train          & 1375             & 765          & 1170    & 1772   \\
\#Val per class  & 20               & 20           & 20      & 150    \\
\#Test per class & 200              & 200          & 200     & 1500   \\
Imbalanced ratio & 17.73            & 5.86         & 9.08    & 4.00      \\ \bottomrule
\end{tabular}
% \vspace{-2em}
\centering\caption{Statistics of datasets. The imbalanced ratio is calculated on the entire graph dataset.}
\label{table:stat}
\end{table}
% \toprule
% Dataset          & \#Nodes & \#Edges & \#Features & \#Classes & \#Train & \#Val per class & \#Test per class & Imbalanced ratio                      \\ \midrule
% Wiki-CS          & 11701   & 216123  & 300        & 10        & 1170   & 20              & 200              & 9.08 \\
% Amazon-Computers & 13752   & 245861  & 767        & 10        & 1375   & 20              & 200              & 17.73 \\
% Amazon-Photo     & 7650    & 119081  & 745        & 8         & 765   & 20              & 200              & 5.86            \\
% DBLP             & 17716   & 105734  & 1639       & 4         & 1772   & 150             & 1500             & 4.00                              \\ \bottomrule

\subsection{Baselines}
The publicly available implementations of graph contrastive learning~(GCL) baselines and their hyperparameters can be found at the following URLs:
\begin{itemize}[leftmargin=10pt]
    \item PyGCL: \url{https://github.com/GraphCL/PyGCL}
    \item DGI: \url{https://github.com/PetarV-/DGI}
    \item MVGRL: \url{https://github.com/kavehhassani/mvgrl}
    \item InfoGraph: \url{https://github.com/fanyun-sun/InfoGraph}
    \item GRACE: \url{https://github.com/CRIPAC-DIG/GRACE}
    \item BGRL: \url{https://github.com/Namkyeong/BGRL_Pytorch}
    \item GBT: \url{https://github.com/pbielak/graph-barlow-twins}
\end{itemize}
\section{Preliminaries of GNN Encoders}
\label{sec:appendix-D}
Typically, the training process of modern GNN encoders follows the message-passing mechanism~\cite{hamilton2020graph} in graph contrastive learning~(GCL). During each message-passing iteration, a hidden embedding $\vh_u^{(k)}$ corresponding to each node $u \in \gV$ is updated by aggregating information from $u$'s neighborhood $\mathcal{N}(u)$, which can be expressed as follows:
\begin{equation}
    \scriptsize
    \begin{aligned}
    \vh_{u}^{(k+1)} &=\text { UPDATE }^{(k)}\left(\vh_{u}^{(k)}, \text { AGGREGATE }^{(k)}\left(\left\{\vh_{v}^{(k)}, \forall v \in \mathcal{N}(u)\right\}\right)\right) \\
    &=\text { UPDATE }^{(k)}\left(\vh_{u}^{(k)}, \vm_{\mathcal{N}(u)}^{(k)}\right),
    \end{aligned}
\end{equation}
where $\text{UPDATE}$ and $\text{AGGREGATE}$ are arbitrary differentiable functions~(\ie neural networks), and $\vm_{\mathcal{N}(u)}$ denotes the ``message'' that is aggregated from $u$'s neighborhood $\mathcal{N}(u)$. The initial embedding at $k=0$ is set to the input features,~\ie $\vh_u^{(0)} = x_u, \forall u \in \mathcal{V}$. After running $k$ iterations of the GNN message-passing, we can obtain information from $k$-hops neighborhood nodes. Different GNNs can be obtained by choosing different $\text{UPDATE}$ and $\text{AGGREGATE}$ functions. For example, the classical Graph Convolutional Networks~(GCN)~\cite{kipf2016semi} updates the hidden embedding as
\begin{equation}
    \mH^{(l+1)}=\sigma\left(\hat{\mA} \mH^{(l)} \mW^{(l)}\right),
\end{equation}
where $\mH^{(l+1)} = \left[\vh_1^{(l+1)};\cdots;\vh_n^{(l+1)}\right]$ is the hidden matrix of the $(l+1)$-th layer. $\hat{\mA}=\hat{\mD}^{-1 / 2}(\mA+\mI) \hat{\mD}^{-1 / 2}$ is the re-normalization of the adjacency matrix, and $\hat{\mD}$ is the corresponding degree matrix of $\mA+\mI$. $\mW^{(l)} \in \sR^{C_l \times C_{l+1}}$ is the filter matrix in the $l$-th layer with $C_l$ referring to the size of $l$-th hidden layer and $\sigma(\cdot)$ is a nonlinear function, \eg ReLU.

\end{document}